\documentclass[10pt,journal,compsoc]{IEEEtran}
\ifCLASSOPTIONcompsoc
  \usepackage[nocompress]{cite}
\else
  \usepackage{cite}
\fi

\usepackage[pdftex]{graphicx}
\usepackage{wrapfig}
\usepackage{xifthen}
\usepackage{xcolor}
\usepackage{colortbl}
\usepackage{setspace}
\usepackage{enumitem}
\usepackage[cmex10]{amsmath}
\usepackage{amssymb}
\usepackage{amsthm}
\usepackage{dsfont}
\usepackage{url}

\usepackage{soul}

\DeclareMathOperator*{\argmax}{arg\,max}

\graphicspath{{images/}}

\begin{document}
\title{Action Understanding\\ with Multiple Classes of Actors}

\author{Chenliang~Xu,~\IEEEmembership{Member,~IEEE,}
        Caiming~Xiong,
        and~Jason~J.~Corso,~\IEEEmembership{Senior Member,~IEEE}% <-this % stops a space
\IEEEcompsocitemizethanks{
\IEEEcompsocthanksitem C. Xu is with the Department of Computer Science, University of Rochester, Rochester, NY 14627. Email: chenliang.xu@rochester.edu.
\IEEEcompsocthanksitem C. Xiong is a senior researcher at Salesforce MetaMind, San Francisco, CA 94105. E-mail: cmxiong.lhi@gmail.com.
\IEEEcompsocthanksitem J. J. Corso is with the Department of Electrical Engineering and Computer Science, University of Michigan, Ann Arbor, MI 48109. E-mail: jjcorso@umich.edu.
}% <-this % stops an unwanted space
%\thanks{Manuscript received April 19, 2005; revised September 17, 2014.}
}


\markboth{Journal of \LaTeX\ Class Files,~Vol.~13, No.~9, September~2014}%
{Shell \MakeLowercase{\textit{et al.}}: Bare Demo of IEEEtran.cls for Computer Society Journals}

%\IEEEpubid{\makebox[\columnwidth]{\hfill 0000--0000/00/\$00.00~\copyright~2014 IEEE}%


\IEEEtitleabstractindextext{%
\begin{abstract}
% Action understanding has received a significant amount of attention in the past decade. Numerous efforts have been made on broadening the scope of actions in consideration, e.g., increased number of action classes and more diverse videos in-the-wild. Despite the progress, existing works focus strictly on one type of action agent, which we call \emph{actor}---a human adult, and ignore the diversity and similarity of actions performed by other actors. 
Despite the rapid progress, existing works on action understanding focus strictly on one type of action agent, which we call \emph{actor}---a human adult, ignoring the diversity of actions performed by other actors. To overcome this narrow viewpoint, our paper marks the first effort in the computer vision community to jointly consider algorithmic understanding of various types of actors undergoing various actions. To begin with, we collect a large annotated Actor-Action Dataset (A2D) that consists of 3782 short videos and 31 temporally untrimmed long videos. We formulate the general actor-action understanding problem and instantiate it at various granularities: video-level single- and multiple-label actor-action recognition, and pixel-level actor-action segmentation. We propose and examine a comprehensive set of graphical models that consider the various types of interplay among actors and actions. Our findings have led us to conclusive evidence that the joint modeling of actor and action improves performance over modeling each of them independently, and further improvement can be obtained by considering the multi-scale natural in video understanding. Hence, our paper concludes the argument of the value of explicit consideration of various actors in comprehensive action understanding and provides a dataset and a benchmark for later works exploring this new problem.
\end{abstract}

}

\maketitle

\IEEEdisplaynontitleabstractindextext

% \fi
\IEEEpeerreviewmaketitle

%--------------------------------------------------------------------
\IEEEraisesectionheading{\section{Introduction}\label{sec:introduction}}

% 1. Action Recognition has seen great progress in the last decade 
\IEEEPARstart{A}{ction} is the heart of video understanding. As such, it has received a significant amount of attention in the last decade. The emphasis has moved from small datasets of a handful of actions~\cite{ScLaCaICPR2004} to large datasets with many dozens of actions~\cite{KuJhGaICCV2011}; from constrained domains like sporting~\cite{NiChFeECCV2010} to videos in-the-wild~\cite{LiLuShCVPR2009}. Notable methods have demonstrated that low-level features~\cite{WaKlScIJCV2013, LaIJCV2005}, mid-level atoms~\cite{ZhWaYaICCV2013}, high-level exemplars~\cite{SaCoCVPR2012}, structured models~\cite{NiChFeECCV2010, TiSuShCVPR2013}, attributes~\cite{LiKuSaCVPR2011}, and even deeply-learned features~\cite{SiZiNIPS2014} can be used for action recognition. Impressive methods have even pushed toward action recognition for multiple views~\cite{MaLaScCVPR2009}, event recognition~\cite{IzShECCV2012}, group-based activities~\cite{LaWaYaNIPS2010}, and even human-object interactions~\cite{GuKeDaTPAMI2009, PeJiZhICCV2011}.

% 2. Argument 1: Existing works in action recognition emphasize only humans. 
However, these many works emphasize a small subset of the broader action understanding problem. First, aside from Iwashita et al.~\cite{IwTaKuICPR2014} who study egocentric animal activities, these existing methods all assume the agent of the action, which we call the \textit{actor}, is a human adult. Although \textit{looking at people} is certainly a relevant application domain for computer vision, it is not the only one; consider recent advances in video-to-text~\cite{XuMeYaCVPR2016, DaXuDoCVPR2013} that can be used for semantic indexing large video databases~\cite{LiFiKoCVPR2014}, or advances in autonomous vehicles~\cite{GeLeUrCVPR2012}. In these applications, understanding both the actor and the action are critical for success: e.g., the autonomous vehicle needs to distinguish between a child, a deer and a squirrel running into the road so it can accurately make an avoidance plan. Applications like these are abundant and growing. Furthermore, if we look to the philosophy of action~\cite{DaBOOK2001, ChXiXuCVPR2014}, we find an axiomatic definition of action: ``first, action is what an agent can do; second, action requires an intention; third, action requires a bodily movement guided by an agent or agents; and fourth, action leads to side-effects''. For comprehensive action understanding, we need to consider not only the action, but also the agent and the movement of the agent. To that end, our recent work has begun to move in this direction. In the visual-psychophysical study~\cite{XuDoHaIJSC2013}, we jointly consider different types of actors performing actions to assess the degree to which a supervoxel segmentation~\cite{XuXiCoECCV2012} retains sufficient semantic information for humans to recognize it. Subsequently, in~\cite{XuHsXiCVPR2015}, we consider a larger problem with seven classes of actors and eight classes of actions and assess the state of the art for joint actor-action understanding; this article is an extended version of this latter work.

% Argument 2: Existing works on restricted action anlaysis. 
Second, in addition to the limited prior work on the different agents of action, the prior literature largely focuses on \textit{action recognition}, which is posed as the classification of a temporally pre-trimmed clip into one of $k$ action classes from a closed-world setting. The direct utility of results based on this problem formulation is limited. The community has indeed begun to move beyond this simplified problem into action detection~\cite{SiMaJoCVPR2016, YeRuMoCVPR2016, PeScECCV2016}, action localization~\cite{SoIdShCVPR2016, ShWaChCVPR2016}, action segmentation~\cite{LeReViECCV2016, XuCoCVPR2016, LuXuCoCVPR2015}, and even actionness ranking~\cite{ChXiXuCVPR2014, LuChTrICCV2015}. Again, all of these works do so strictly in the context of human actors and furthermore, many assume that videos are temporally pre-trimmed to contain only single type of action. 

\begin{figure*}[t]
\centering
\includegraphics[width=1\linewidth]{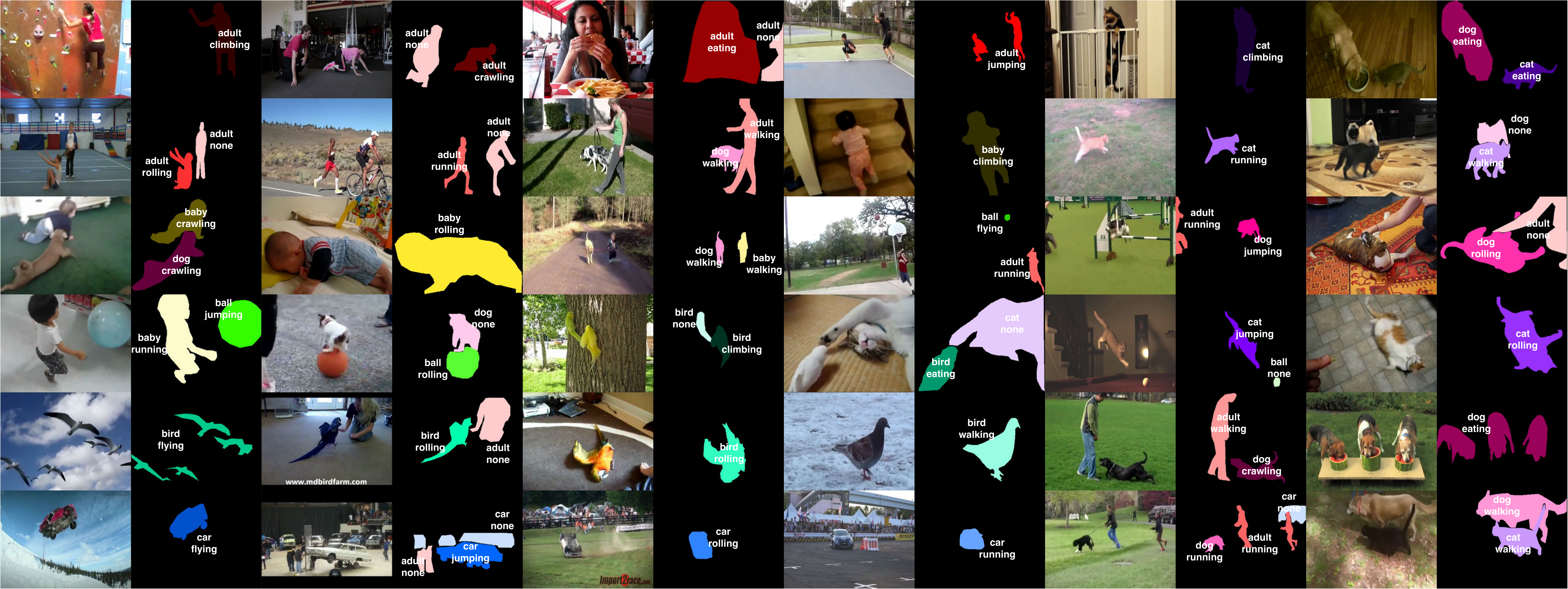}
\caption{Labeled frames from A2D short videos. Colors are picked from the HSV color space, so that the same actors have the same \textit{hue} (refer to Fig. \ref{fig:dat:color} for the color-legend). Black is the background. Best view zoomed and in color.}
\label{fig:img:short}
\end{figure*}

\begin{figure*}[t]
\centering
\includegraphics[width=1\linewidth]{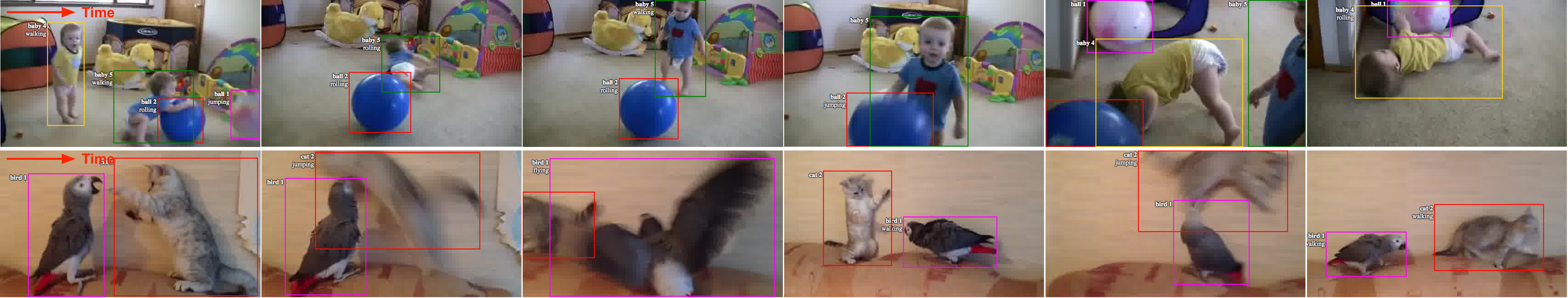}
\caption{Sampled frames from A2D long videos. Actors are annotated with tracked bounding boxes over time, and their actions are shown next to the bounding boxes. Best view zoomed and in color.}
\label{fig:img:long}
\end{figure*}

% Our Paper: 1) The Problem.
In this paper, we overcome both of these narrow viewpoints and introduce a new level of generality to the action understanding problem by considering multiple different classes of actors undergoing multiple different classes of actions. To be exact, we consider seven actor classes (\textit{adult}, \textit{baby}, \textit{ball}, \textit{bird}, \textit{car}, \textit{cat}, and \textit{dog}) and eight action classes (\textit{climb}, \textit{crawl}, \textit{eat}, \textit{fly}, \textit{jump}, \textit{roll}, \textit{run}, and \textit{walk}) not including the no-action class, which we also consider. We formulate a general actor-action understanding framework and implement it for three specific problems: actor-action recognition with single- and multiple-label; and actor-action segmentation. These three problems cover different levels of modeling and hence allow us to analyze the new problem thoroughly. 

% Our Paper:  2) The Dataset.
To support these new actor-action understanding problems, we have created a new dataset, which we call the Actor-Action Dataset or A2D. The A2D has 3782 short videos with at least 99 instances per valid actor-action tuple and 31 long videos for a selected subset of actor-action tuples. The short videos are temporally trimmed to a few seconds such that the individual actors perform only one type of action in each video, and the videos are labeled at the pixel-level for actors and actions (densely in space over actors, sparsely in time, see Fig.~\ref{fig:img:short}). The long videos range from a half-minute to two minutes and the individual actors may perform multiple actions over time, where action-tagged bounding boxes are annotated on the actors (densely in time, see Fig.~\ref{fig:img:long}). 

% Our Paper: 3) The Models.
We thoroughly analyze empirical performance of both baseline and state-of-the-art graphical models in the context of modeling the interplay of actors and actions. The baseline models include a na\"ive Bayes model (independent over actors and actions) and a joint product-space model (each actor-action pair is considered as one class). We adapt the bilayer model from~\cite{LaStRuIJCV2012} that considers the compatibilities of actors performing actions; we propose a trilayer model, where two sets of conditional classifier-based edges are added to connect  the bilayer and the joint product-space. These edges explore even finer detail, i.e., the diversity and similarity of various actions performed by various actors. The above models are evaluated at both video-level recognition tasks and pixel-level segmentation task. Observing the performance gap between video-level and pixel-level tasks (the latter is harder), we further explore a multi-scale approach to incorporate video-level recognition responses in optimizing pixel-level actor-action segmentation. 

% Our Conclusion ...
Our experiments demonstrate that inference jointly over actors and actions outperforms inference independently over them, and hence, supports the explicit consideration of various actors in comprehensive action understanding. In other words, although a \emph{bird} and an \emph{adult} can both \emph{eat}, the space-time appearance of a \emph{bird eating} and an \emph{adult eating} are different in significant ways. Furthermore, the various mannerisms of the way \emph{birds eat} and \emph{adults eat} mutually reinforces inference over the constituent parts. This result is analogous to Sadeghi and Farhadi's visual phrases work~\cite{SaFaCVPR2011} in which it is demonstrated that joint detection over small groups of objects in images is more robust than separate detection over each object followed a merging process, and to Gupta et al.'s~\cite{GuKeDaTPAMI2009} work on human object-interactions in which considering specific objects while modeling human actions leads to better inferences for both parts. A secondary founding is that the performances of all models considered in actor-action segmentation get improved when both pixel-level and video-level evidences are considered, which further suggests the necessity of multi-scale modeling in understanding videos. 

% Contributions & What's new in Journal?
Our paper presents the first effort in the computer vision community to jointly consider various types of actors undergoing various actions.  As such, we pose three goals: 
\begin{enumerate}
\item we seek to formulate the general actor-action understanding problem and instantiate it at various granularities; 
\item we assess whether it is beneficial to explicitly jointly consider actors and actions in this new problem-space using both short and long videos; 
\item we explore a multi-scale approach that bridges the gap between recognition and  segmentation tasks. 
\end{enumerate}
Notice that an early version of our work appears in~\cite{XuHsXiCVPR2015}. In this journal article, we make the following additions: 1) we extend the A2D dataset with 31 long videos, a total of 40 minutes, that contain lots of actor-action changes in time; 2) we conduct experiments on long videos using newly devised metric to evaluate temporal performance of segmentation models; 3) we apply a new inference scheme that achieves 10 times faster inference than the conference paper;  4) we explore a multi-scale approach to incorporate video-level recognition responses via label costs in inferring pixel-level segmentation; and 5) we provide more detail of our methods and a new related work section. 

% Paper organize
The rest of the paper is organized as follows: Sec.~\ref{sec:related} discusses related work; Sec.~\ref{sec:data} introduces the extended A2D dataset; Sec.~\ref{sec:prob} formulates the general actor-action understanding problem and describes various models to consider actor-action interactions; Sec.~\ref{sec:holistic} describes a multi-scale approach to actor-action segmentation; Sec.~\ref{sec:exp:short} and Sec.~\ref{sec:exp:long} detail the experiments conducted on short videos and long videos, respectively; and Sec.~\ref{sec:con} concludes our paper. 

%--------------------------------------------------------------------
\section{Related Work}
\label{sec:related}

Our work explores a new dimension in fine-grained action understanding. A related work is Bojanowski et al.~\cite{BoBaLaICCV2013}, where they focus on finding different human \textit{actors} in movies, but these are the actor-names and not different types of actors, like \textit{dog} and \textit{cat} as we consider in this paper. In~\cite{IwTaKuICPR2014}, Iwashita et al. study egocentric animal activities, such as turn head, walk and body shake, from a view-point of an animal---a dog. Our work differs from them by explicitly considering the types of actors in modeling actions. Similarly, our work also differs from the existing works on actions and objects, such as \cite{GuKeDaTPAMI2009, PeJiZhICCV2011}, which are strictly focused on interactions between human actors manipulating various physical objects.

In addition to the increased diversity in acitivities, the literature in action understanding is moving away from simple video classification to fine-grained output. For example, methods like~\cite{TiSuShCVPR2013, PeScECCV2016} detect human actions with bounding box tubes or 3D volumes in videos; and methods like~\cite{LuXuCoCVPR2015, SoIdShCVPR2016} even provide pixel-level segmentation of human actions. The shift of research interest is also observed in the relaxed assumptions of videos. For instance, temporally untrimmed videos are considered in~\cite{ShWaChCVPR2016} and~\cite{YuNiYaCVPR2016}, and online streaming videos are explored in~\cite{SoIdShCVPR2016} and~\cite{GeGaGhECCV2016}. Our work analyzes actors and actions at various granularities, e.g., both video-level and pixel-level, and we experiment with both short and long videos. 

%--------------------------------------------------------------------
\subsection{From Segmentation Perspective}

We now discuss related work in segmentation, which is a major emphasis of our broader view of  action understanding, e.g., the actor-action segmentation task. Semantic segmentation methods can now densely label more than a dozen classes in images~\cite{LaRuKoICCV2009, MoChLiCVPR2014} and videos~\cite{TiLaIJCV2012, JaChViICCV2013} undergoing rapid motion; semantic segmentation methods have even been unified with object detectors and scene classifiers~\cite{YaFiUrCVPR2012}, extended to 3D~\cite{KuLiDaECCV2014} and posed jointly with attributes~\cite{ZhChWaCVPR2014}, stereo~\cite{LaStRuIJCV2012} and SFM~\cite{BrShFaECCV2008}. Although the underlying optimization problems in these methods tend to be expensive, average per-class accuracy scores have significantly increased, e.g., from 67\% in~\cite{ShWiRoIJCV2009} to nearly 80\% in~\cite{LaRuKoICCV2009, KrKoNIPS2011, YaFiUrCVPR2012} on the popular MSRC semantic segmentation benchmark. Further works have moved beyond full supervision to weakly supervised object discovery and learning~\cite{TaSuYaCVPR2013, XuScUrCVPR2015}. However, these existing works in semantic segmentation focus on labeling pixels/voxels as various objects or background-stuff classes, and they do not consider the joint label-space of what actions these ``objects'' may be doing. Our work differs from them by directly considering this actor-action problem, while also building on the various advances made in these papers. 

Other related works include video object segmentation~\cite{LeKiGrICCV2011, LiKiHuICCV2013} and joint temporal segmentation with action recognition~\cite{HoLaDeCVPR2011}. The video object segmentation methods are class-independent and assume a single dominant object (actor) in the video; they are hence not directly comparable to our work although one can foresee a potential method using video object segmentation as a precursor to the actor-action understanding problem. 

%--------------------------------------------------------------------
\subsection{From Model Perspective}

Multi-task learning has been effective in many applications, such as object detection~\cite{SaToTeCVPR2011}, and classification~\cite{YuLiYaTIP2012}. The goal is to learn models or shared representations jointly that outperforms learning them separately for each task. Recently, multi-task learning has also been adapted to action classification. For example, in~\cite{ZhWaJiICCV2013}, Zhou et al. classify human actions in videos with shared latent tasks. Our paper differs from them by explicitly modeling the relationship and interactions among  actors and actions under a unified graphical model. Notice that it is possible to use multi-task learning to train a shared deep representation for actors and actions extending frameworks such as the multi-task network cascades~\cite{DaHeSuCVPR2016}, which has a different focus than this work; hence, we leave it to future work.

%--------------------------------------------------------------------
\section{A2D v2---The Extended Actor-Action Dataset}
\label{sec:data}

\begin{figure}[t]
\centering
\includegraphics[width=0.99\linewidth]{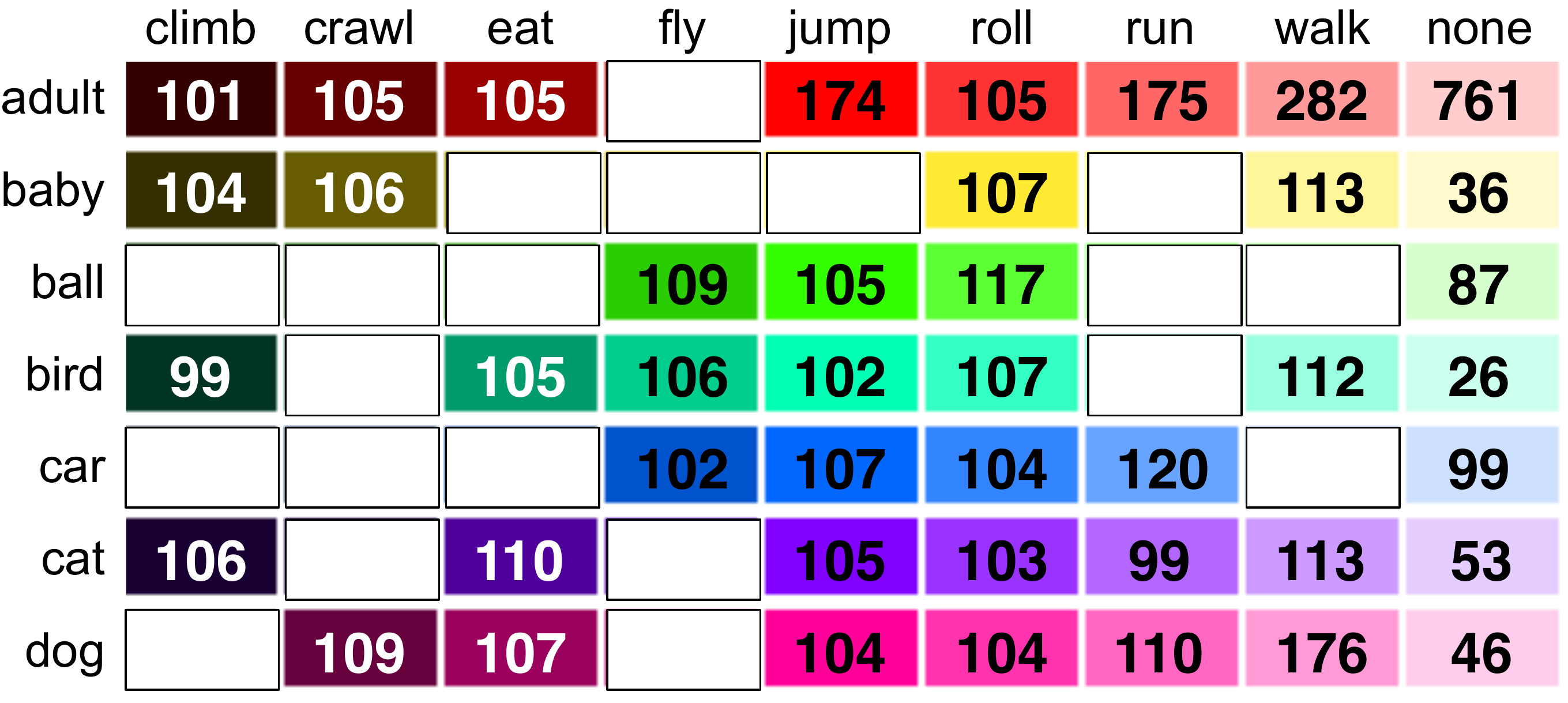}
\caption{Statistics of short videos containing a given actor-action label. Empty entries are invalid labels or labels in insufficient supply. The background color depicts the color we use for the label throughout the paper; we vary hue for actor and saturation for action.}
\label{fig:dat:color}
\end{figure}

The A2D v2 has 3782 short videos from the initial A2D~\cite{XuHsXiCVPR2015} and 31 new long videos. All videos are collected from YouTube; and they are hence unconstrained ``in-the-wild'' videos with varying characteristics. We select seven classes of actors performing eight different actions. Our choice of actors covers articulated ones, such as \textit{adult}, \textit{baby}, \textit{bird}, \textit{cat} and \textit{dog}, as well as rigid ones, such as \textit{ball} and \textit{car}. The eight actions are \textit{climbing}, \textit{crawling}, \textit{eating}, \textit{flying}, \textit{jumping}, \textit{rolling}, \textit{running}, and \textit{walking}. A single action-class can be performed by mutiple actors, but none of the actors can perform all eight actions.  For example, we do not consider \emph{adult-flying} or \emph{ball-running} in the dataset. In some cases, we have pushed the semantics of the given action term to maintain a small set of actions: e.g., \textit{car-running} means the car is moving and \textit{ball-jumping} means the ball is bouncing. One additional action label \textit{none} is added to account for actions other than the eight listed ones and actors that are not performing an action. Therefore, we have in total 43 valid actor-action tuples (see Fig. \ref{fig:dat:color} colored entries). This allows us to explore the valid combinations of actor-action tuples, rather than brute-force enumeration.

%--------------------------------------------------------------------
\subsection{Short Videos}
\label{sec:dat:short}

\begin{figure}[t]
\centering
\includegraphics[width=0.80\linewidth]{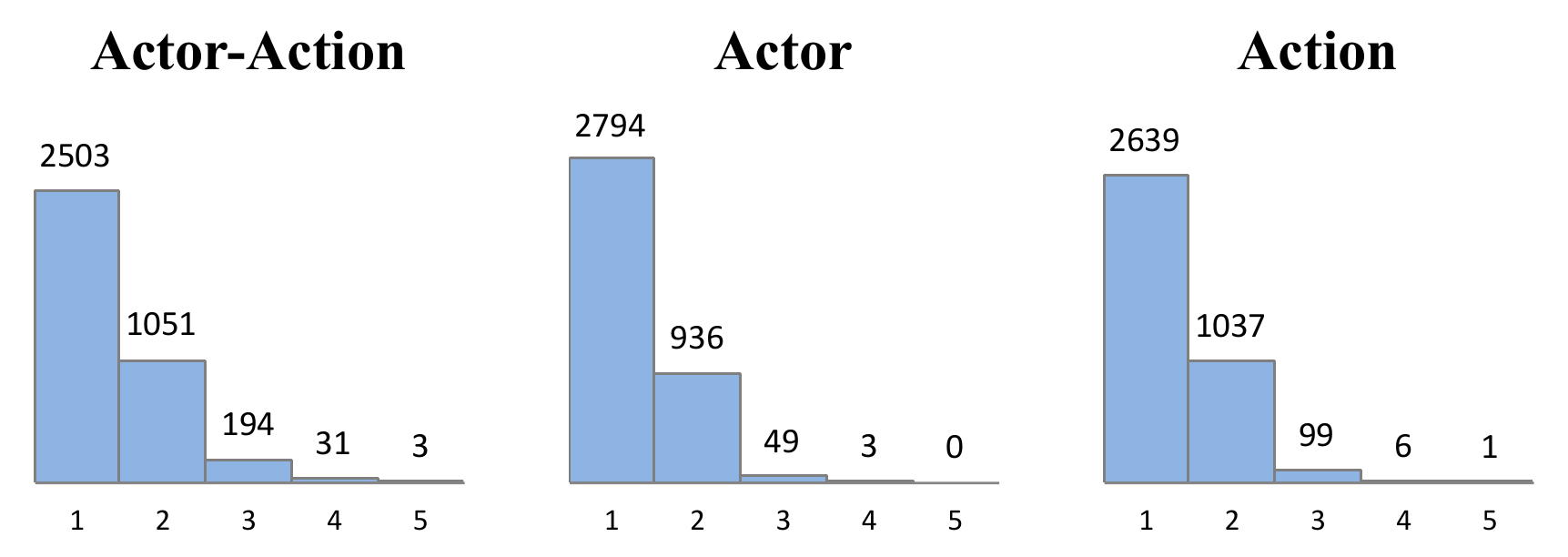}
\caption{Histograms of short videos w.r.t. the number of concurrent of actor-action, actor and action labels. Roughly one-third of the short videos have more than one actor and/or action.}
\label{fig:hist:short}
\end{figure}

We use various text-searches generated from actor-action tuples to query the YouTube database.  Resulting videos are then manually trimmed to contain at least one instance of the primary actor-action tuple, and they are also encouraged to contain optionally many actor-action tuples. The trimmed short videos have an average length of 136 frames, with a minimum of 24 frames and a maximum of 332 frames. We split them into 3036 training videos and 746 testing videos divided evenly over all actor-action tuples. Figure~\ref{fig:dat:color} (numbers) shows the counts of short videos for each actor-action tuple. Notice that one-third of the short videos have more than one actor performing different actions (see Fig.~\ref{fig:hist:short} for exact counts), which further distinguishes our dataset from existing action classification datasets, e.g.,~\cite{KuJhGaICCV2011, SoZaShARXIV2012}. 

To support the broader set of action understanding problems in consideration, we label three to five frames for each short video with dense pixel-level actor-action labels. Figure~\ref{fig:img:short} shows sampled frame examples. The selected frames are evenly distributed over a video. We start by collecting crowd-sourced annotations from MTurk using the LabelMe toolbox~\cite{RuToMuIJCV2008}, then we manually filter each video to ensure the labeling quality as well as the temporal coherence of labels. In total, we have collected 11936 labeled frames. Video-level labels are computed directly from these pixel-level labels for the recognition tasks. To the best of our knowledge, our dataset is the first video dataset that contains both pixel-level actor and action annotations.  

%--------------------------------------------------------------------
\subsection{Long Videos}
\label{sec:dat:long}

\begin{figure}[t]
\centering
\includegraphics[width=0.70\linewidth]{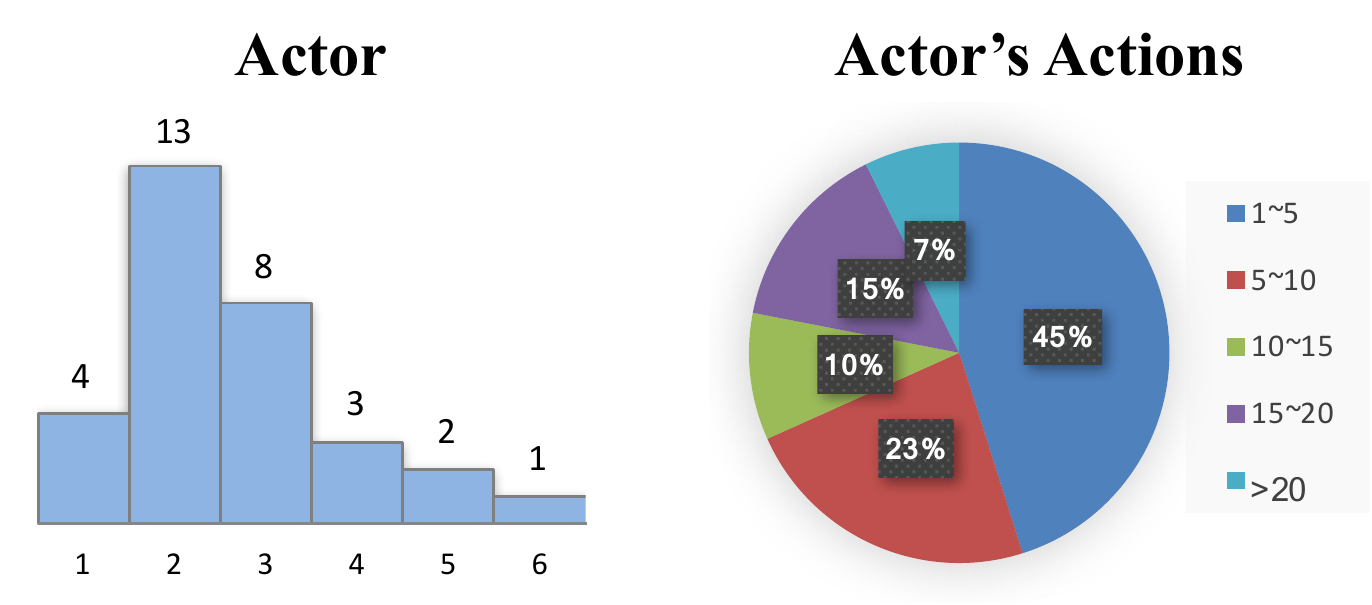}
\caption{Left: Histogram of long videos w.r.t. the number of concurrent actors. Right: Pie chart of actors w.r.t. their performed number of actions in a long video.}
\label{fig:hist:long}
\end{figure}

We add 31 temporally untrimmed long videos to A2D~v2, where a single actor has many changes of actions over time in a video. These videos are more realistic comparing to manually trimmed short videos. The long videos also contain multiple actors, thus they are very different than the untrimmed videos used in existing action localization challenges such as the ActivityNet~\cite{CaEsGhCVPR2015}, which allows us to analyze the actor-action problems in both space and time.

The long videos have an average length of 1888 frames, with a minimum of 593 frames and a maximum of 3605 frames. The actors in long videos may perform a series of actions. For example, a \textit{dog} may perform \textit{walking}, then \textit{running}, then \textit{walking} in a video. The statistics are shown in Fig.~\ref{fig:hist:long}. For long videos, we reserve 8 videos for validation and use the remaining 23 videos for testing. We use VATIC annotation tool~\cite{VoPaRaIJCV2012} to track actors with bounding boxes through a video, and label actions densely in time. Figure~\ref{fig:img:long} shows examples on two long videos.

%--------------------------------------------------------------------
\section{Actor-Action Understanding Problems}
\label{sec:prob}

Without loss of generality, let $\mathcal{V} = \{v_1, \dots, v_n\}$ denote a video with $n$ voxels in space-time lattice $\Lambda^3$ or $n$ supervoxels in a video segmentation~\cite{XuCoCVPR2012, XuXiCoECCV2012, ChWeIICVPR2013} represented as a graph $\mathcal{G} = (\mathcal{V},\mathcal{E})$, where the neighborhood structure is given by the supervoxel segmentation method; when necessary we write $\mathcal{E}(v)$, where $v \in \mathcal{V}$, to denote the subset of $\mathcal{V}$ that are neighbors with $v$. We use $\mathcal{X}$ to denote the set of actor labels: $\{$\textit{adult}, \textit{baby}, \textit{ball}, \textit{bird}, \textit{car}, \textit{cat}, \textit{dog}$\}$, and we use $\mathcal{Y}$ to denote the set of action labels: $\{$\textit{climbing}, \textit{crawling}, \textit{eating}, \textit{flying}, \textit{jumping}, \textit{rolling}, \textit{running}, \textit{walking}, \textit{none}\footnote{The \textit{none} action means either there is no action present or the action is not one of those we have considered.}$\}$. 

Consider a set of random variables $\mathbf{x}$ for actor and another $\mathbf{y}$ for action; the specific dimensionality of $\mathbf{x}$ and $\mathbf{y}$ will be defined later.  Then, the general actor-action understanding problem is specified as a posterior maximization:
\begin{align}
(\mathbf{x}^*,\mathbf{y}^*) = \argmax_{\mathbf{x},\mathbf{y}} 
P(\mathbf{x},\mathbf{y}|\mathcal{V})
\enspace.
\label{eq:obj}
\end{align}
Specific instantiations of this optimization problem give rise to various actor-action understanding problems, which we specify next, and specific models for a given instantiation will vary the underlying relationship between $\mathbf{x}$ and $\mathbf{y}$ allowing us to deeply understand their interplay.

%--------------------------------------------------------------------
\subsection{Single-Label Actor-Action Recognition}  
\label{sec:prob:single}

This is the coarsest level of granularity we consider and it instantiates the standard action recognition problem~\cite{LaIJCV2005}. Here, $\mathbf{x}$ and $\mathbf{y}$ are simply scalars $x$ and $y$, respectively, depicting the single actor and action label to be specified for a given video $\mathcal{V}$. We consider three models for this case:
\\
\textbf{Na\"ive Bayes-Based:} Assume independence across actions and actors, and then train a set of classifiers over actor space $\mathcal{X}$ and a separate set of classifiers over action space $\mathcal{Y}$. This is the simplest approach and is not able to enforce actor-action tuple 
existence: e.g., it may infer invalid \textit{adult-flying}.
\\
\textbf{Joint Product Space:} Create a new label space $\mathcal{Z}$ that is the joint product space of actors and actions: $\mathcal{Z} = \mathcal{X} \times \mathcal{Y}$.  Directly learn a classifier for each actor-action tuple in this joint product space. Clearly, this approach enforces actor-action tuple existence, and we expect it to be able to exploit cross-actor-action features to learn more discriminative classifiers. However, it may not be able to exploit the commonality across different actors or actions, such as the similar manner in which a \textit{dog} and a \textit{cat} \textit{walk}.
\\
\textbf{Trilayer:} The trilayer model unifies the na\"ive Bayes and the joint product space models. It learns classifiers over the actor space $\mathcal{X}$, the action space $\mathcal{Y}$ and the joint actor-action space $\mathcal{Z}$. During inference, it separately infers the na\"ive Bayes terms and the joint product space terms and then takes a linear combination of them to yield the final score. It models not only the cross-actor-action but also the common characteristics among the same actor performing different actions as well as the different actors performing the same action. 

In all cases, we extract local features (see Sec.~\ref{sec:exp:single} for details) and train a set of one-versus-all classifiers, as is standard in contemporary action recognition methods, and although not strictly probabilistic, can be interpreted as such to implement Eq.~\ref{eq:obj}.

%--------------------------------------------------------------------
\subsection{Multiple-Label Actor-Action Recognition}
\label{sec:prob:multi}

As noted in Fig.~\ref{fig:hist:short}, about one-third of the short videos in A2D v2 have more than one actor and/or action present in a given video. In many realistic video understanding applications, we find such multiple-label cases. We address this explicitly by instantiating Eq.~\ref{eq:obj} for the multiple-label case. Here, $\mathbf{x}$ and $\mathbf{y}$ are binary vectors of dimension $|\mathcal{X}|$ and $\lvert\mathcal{Y}\rvert$ respectively. The $x_i$ takes value one if the $i$th actor-type is present in the video and zero otherwise. We define $\mathbf{y}$ similarly. This general definition does not tie specific elements of $\mathbf{x}$ to those in $\mathbf{y}$. Therefore, it allows us to compare independent multiple-label performance over actors and actions within that of the actor-action tuples. We again consider a na\"ive Bayes pair of multiple-label actor and action classifiers, multiple-label actor-action classifiers over the joint product space, as well as the trilayer model that unifies the above classifiers.

%--------------------------------------------------------------------
\subsection{Actor-Action Segmentation}
\label{sec:prob:seg}

Actor-action segmentation is the most fine-grained instantiation that we consider, and it subsumes other coarser problems like detection and localization, which we do not consider in this paper. Here, we seek a label for actor and action per-voxel over the entire video. We define two sets of random variables $\mathbf{x} = \{x_1, \dots, x_n\}$ and $\mathbf{y} = \{y_1, \dots, y_n\}$ to have dimensionality in the number of voxels or supervoxels, and assign each $x_i \in \mathcal{X}$ and each $y_i \in \mathcal{Y}$.  The objective function in Eq.~\ref{eq:obj} remains the same, but the way we define the graphical model implementing $P(\mathbf{x},\mathbf{y}|\mathcal{V})$ leads to acutely different assumptions on the relationship between actor and action variables.

\begin{figure}[t]
\centering
\includegraphics[width=1.0\linewidth]{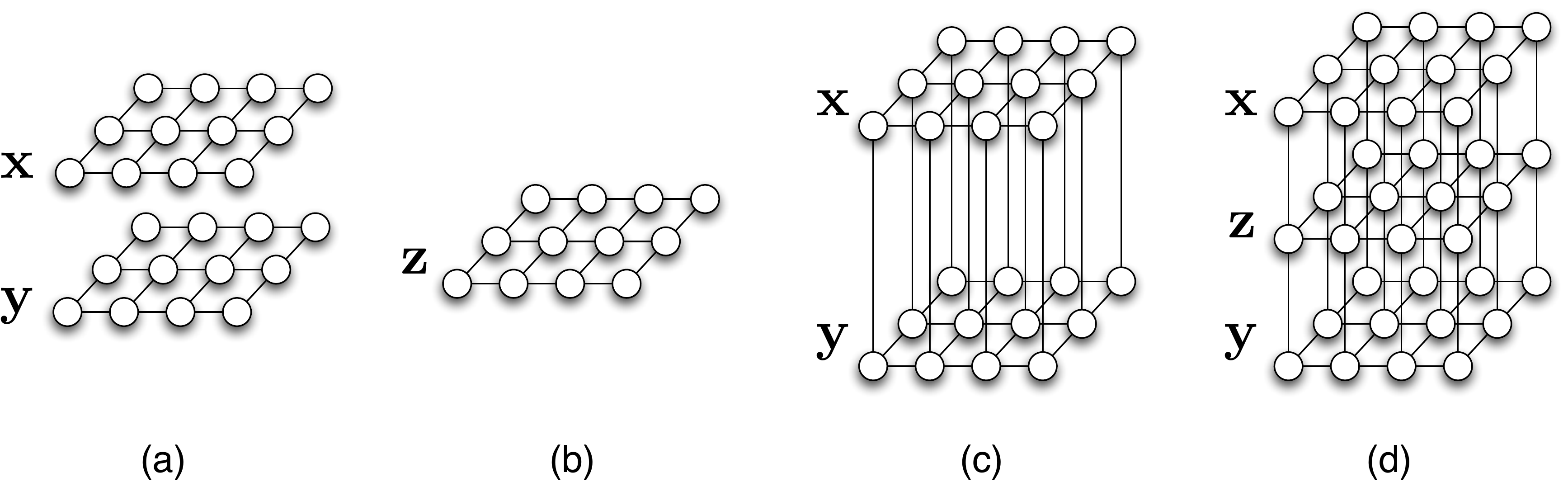}
\caption{Visualization of different graphical models to solve Eq.~\ref{eq:obj}. The figure here is for simple illustration and the actual voxel or supervoxel graph is built for a video volume.}
\label{fig:models}
\end{figure}

We explore this relationship in the remainder of this section. We start by again introducing a na\"ive Bayes-based model in Sec.~\ref{sec:prob:seg:nb} that treats the two classes of labels separately, and a joint product space model in Sec.~\ref{sec:prob:seg:jp} that considers actors and actions together in a tuple $[\mathbf{x}, \mathbf{y}]$. We then explore a bilayer model in Sec.~\ref{sec:prob:seg:bi}, inspired by  Ladick\'{y} et al.~\cite{LaStRuIJCV2012}, that considers the inter-set relationship between actor and action variables. Following that, we introduce a trilayer model in Sec.~\ref{sec:prob:seg:tri} that considers both intra- and inter-set relationships. Figure~\ref{fig:models} illustrates these various graphical models. Finally, we show that these models can be efficiently solved by graph cuts inference in Sec.~\ref{sec:prob:seg:infer}. 

%--------------------------------------------------------------------
\subsubsection{Na\"ive Bayes-based Model}
\label{sec:prob:seg:nb}

First, let us consider a na\"ive Bayes-based model, similar to the one used for actor-action recognition earlier: 
\begin{align}
&P(\mathbf{x},\mathbf{y}|\mathcal{V})
= P(\mathbf{x}|\mathcal{V})P(\mathbf{y}|\mathcal{V})
\label{eq:nb}
\\
\begin{split}
&= \prod_{i\in\mathcal{V}} P(x_i)P(y_i) \prod_{i\in\mathcal{V}} 
\prod_{j\in\mathcal{E}(i)} P(x_i,x_j)P(y_i,y_j)
\nonumber\\
&\propto \prod_{i\in\mathcal{V}} \phi_i (x_i) \psi_i (y_i)
\prod_{i\in\mathcal{V}} \prod_{j\in\mathcal{E}(i)} \phi_{ij} (x_i,x_j) 
\psi_{ij}(y_i,y_j)
\end{split}
\nonumber
\end{align}
where $\phi_i$ and $\psi_i$ encode the separate potential functions defined on actor and action nodes alone, respectively, and $\phi_{ij}$ and $\psi_{ij}$ are the pairwise potential functions within sets of actor nodes and sets of action nodes, respectively. 

We train classifiers $\{f_c|c \in \mathcal{X}\}$ over actors and $\{g_c|c \in \mathcal{Y}\}$ on sets of actions using features described in Sec.~\ref{sec:exp:seg}, and $\phi_i$ and $\psi_i$ are the classification scores for supervoxel $i$. The pairwise edge potentials have the form of a contrast-sensitive Potts model~\cite{BoJoICCV2001}:
\begin{align}
\phi_{ij} = \left \{\begin{array}{ll}
1 & \mbox{if $x_i=x_j$ }\\
\exp(-\theta/(1+\chi_{ij}^2)) & \text{otherwise},\end{array}
\right.
\label{eq:pairwise}
\end{align}
where $\chi_{ij}^2$ is the $\chi^2$ distance between feature histograms of nodes $i$ and $j$, $\theta$ is a parameter to be learned from the training data. $\psi_{ij}$ is defined analogously.  Actor-action semantic segmentation is obtained by solving these two \textit{flat} CRFs independently.

%--------------------------------------------------------------------
\subsubsection{Joint Product Space}
\label{sec:prob:seg:jp}

We consider a new set of random variables $\mathbf{z} = \{z_1, \dots, z_n\}$ defined again on all supervoxels in a video and take labels from the actor-action product space $\mathcal{Z} = \mathcal{X}\times\mathcal{Y}$. This formulation jointly captures the actor-action tuples as unique entities but cannot model the common actor and action behaviors among different tuples as later models below do; we hence have a single-layer graphical model:
\begin{align}
&P(\mathbf{x},\mathbf{y}|\mathcal{V})
\doteq P(\mathbf{z}|\mathcal{V}) = \prod_{i\in\mathcal{V}} P(z_i) 
\prod_{i\in\mathcal{V}} \prod_{j\in\mathcal{E}(i)} P(z_i,z_j)
\quad\quad
\nonumber\\
&\quad\propto \prod_{i\in\mathcal{V}} \varphi_i(z_i)
\prod_{i\in\mathcal{V}} \prod_{j\in\mathcal{E}(i)} \varphi_{ij} (z_i, z_j)
\label{eq:jp}\\
&\quad= \prod_{i\in\mathcal{V}} \varphi_i([x_i,y_i])
\prod_{i\in\mathcal{V}} \prod_{j\in\mathcal{E}(i)} \varphi_{ij} ([x_i,y_i], [x_j,y_j])
\nonumber
\enspace,
\end{align}
where $\varphi_i$ is the potential function for joint actor-action product space label, and $\varphi_{ij}$ is the inter-node potential function between nodes with the tuple $[\mathbf{x}, \mathbf{y}]$. To be specific, $\varphi_i$ contains the classification scores on the node $i$ from running trained actor-action classifiers $\{h_c|c\in\mathcal{Z}\}$, and $\varphi_{ij}$ has the same form as Eq.~\ref{eq:pairwise}. Figure~\ref{fig:models} (b) illustrates this model as a one layer CRF defined on the actor-action product space.

%--------------------------------------------------------------------
\subsubsection{Bilayer Model}
\label{sec:prob:seg:bi}

Given the actor nodes $\mathbf{x}$ and action nodes~$\mathbf{y}$, the bilayer model connects each pair of random variables $\{(x_i, y_i)\}_{i=1}^n$ with an edge that encodes the potential function for the tuple $[x_i, y_i]$, directly capturing the \textit{covariance} across the actor and action labels. Therefore, we have:
\begin{align}
P(\mathbf{x},\mathbf{y}|\mathcal{V}) = &
P(\mathbf{x}|\mathcal{V}) P(\mathbf{y}|\mathcal{V}) 
\prod_{i\in\mathcal{V}} P(x_i,y_i)
\nonumber\\
\propto &\prod_{i\in\mathcal{V}} \phi_i(x_i) \psi_i(y_i) \xi_i(x_i,y_i)
\cdot\nonumber\\
&\prod_{i\in\mathcal{V}} \prod_{j\in\mathcal{E}(i)} \phi_{ij}(x_i,x_j) 
\psi_{ij}(y_i,y_j)
\enspace,
\label{eq:bi}
\end{align}
where $\phi_\cdot$ and $\psi_\cdot$ are defined as earlier, $\xi_i(x_i,y_i)$ is a learned potential function over the product space of labels, which can be exactly the same as $\varphi_i$ in Eq.~\ref{eq:jp} above or a compatibility term like the contrast sensitive Potts model, Eq.~\ref{eq:pairwise} above. We choose the former in this paper. Fig.~\ref{fig:models}~(c) illustrates this model. We note that additional links can be constructed by connecting corresponding edges between neighboring nodes across layers and encoding the occurrence among the bilayer edges, such as the joint object class segmentation and dense stereo reconstruction model in Ladick\'{y} et al.~\cite{LaStRuIJCV2012}. However, their model is not directly suitable here.

%--------------------------------------------------------------------
\subsubsection{Trilayer Model}
\label{sec:prob:seg:tri}

So far we have introduced three baseline formulations of Eq.~\ref{eq:obj} for actor-action segmentation that relate the actor and action terms in different ways. The na\"ive Bayes model (Eq. \ref{eq:nb}) does not consider any relationship between actor $\mathbf{x}$ and action $\mathbf{y}$ variables. The joint product space model (Eq. \ref{eq:jp}) combines features across actors and actions as well as inter-node interactions in the neighborhood of an actor-action node. The bilayer model (Eq. \ref{eq:bi}) adds actor-action interactions among separate actor and action nodes, but it does not consider how these interactions vary spatiotemporally. 

Therefore, we introduce a new trilayer model that explicitly models such variations (see Fig.~\ref{fig:models} (d)) by combining nodes $\mathbf{x}$ and $\mathbf{y}$ with the joint product space nodes $\mathbf{z}$: 
\begin{align} 
P(\mathbf{x},\mathbf{y},&\mathbf{z}|\mathcal{V}) = P(\mathbf{x}|\mathcal{V}) 
P(\mathbf{y}|\mathcal{V}) P(\mathbf{z}|\mathcal{V}) \prod_{i\in\mathcal{V}} 
P(x_i,z_i) P(y_i,z_i)\nonumber\\
\propto &
\prod_{i\in\mathcal{V}} \phi_i(x_i) \psi_i(y_i) \varphi_i(z_i)
\mu_i(x_i,z_i) \nu_i(y_i,z_i)\cdot\nonumber\\
&
\prod_{i\in\mathcal{V}} \prod_{j\in\mathcal{E}(i)} \phi_{ij}(x_i,x_j) 
\psi_{ij}(y_i,y_j) \varphi_{ij}(z_i, z_j)
\enspace,
\label{eq:tri}
\end{align}
where we further define:
\begin{align}
\mu_i(x_i,z_i)=\left \{\begin{array}{ll}
w({y_i}'|x_i) & \mbox{if $x_i={x_i}'$ for $z_i=[{x_i}',{y_i}']$ }\\
0 & \text{otherwise}\end{array}
\right.
\label{eq:tri_w}
\\
\nu_i(y_i,z_i)=\left \{\begin{array}{ll}
w( {x_i}'|y_i) & \mbox{if $y_i={y_i}'$ for $z_i=[{x_i}',{y_i}']$ }\\
0 & \text{otherwise}\end{array}.
\right.
\nonumber
\end{align}
Terms $w(y_i'|x_i)$ and $w(x_i'|y_i)$ are classification scores of conditional classifiers, which are explicitly trained for this trilayer model. These conditional classifiers are the main reason for the increased performance found in this method: separate classifiers for the same action conditioned on the type of actor are able to exploit the characteristics unique to that actor-action tuple. For example, when we train a conditional classifier for action \textit{eating} given actor \textit{adult}, we use all other actions performed by \textit{adult} as negative training samples. Therefore our trilayer model considers all relationships in the individual actor and action spaces as well as the joint product space. In other words, the previous three baseline models are all special cases of the trilayer model.  It can be shown that the solution $(\mathbf{x}^*,\mathbf{y}^*,\mathbf{z}^*)$ maximizing Eq.~\ref{eq:tri} also maximizes Eq.~\ref{eq:obj} (see Appendix).

%--------------------------------------------------------------------
\subsubsection{Inference}
\label{sec:prob:seg:infer}

\begin{figure}[t]
\centering
\includegraphics[width=0.80\linewidth]{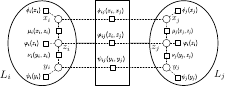}
\caption{Visualization of two nodes in the trilayer model after rewritten. The dashed lines are edges before rewritten. The bold ellipses are new nodes and the solid bold line is the new edge.}
\label{fig:inference}
\end{figure}

Until now we have described all models solving Eq.~\ref{eq:obj}, and they can be optimized using loopy belief propagation in our previous paper~\cite{XuHsXiCVPR2015}. In this section, we show that they can also be optimized using the graph cuts~\cite{BoVeZaTPAMI2001, BoKoTPAMI2004, KoZaTPAMI2004}, which greatly improves the inference efficiency. We use the trilayer model here as an illustration example and we note that the bilayer model can be solved in a similar way. 

We define a new set of random variables $\mathbf{L} = \{(L^o,L^a)_1, \dots, (L^o,L^a)_n | L^o \in \mathcal{X}, L^a \in \mathcal{Y}\}$ on voxels in a video or supervoxels of a video segmentation. Therefore, we can rewrite Eq.~\ref{eq:tri} in the following form:
\begin{align}
&P(\mathbf{x},\mathbf{y},\mathbf{z}|\mathcal{V}) 
\doteq P(\mathbf{L}|\mathcal{V}) = \prod_{i\in\mathcal{V}} P(L_i) 
\prod_{i\in\mathcal{V}} \prod_{j\in\mathcal{E}(i)} P(L_i,L_j)
\quad\quad
\nonumber\\
&\quad\propto \prod_{i\in\mathcal{V}} \Phi_i(L_i)
\prod_{i\in\mathcal{V}} \prod_{j\in\mathcal{E}(i)} \Phi_{ij} (L_i, L_j)
\enspace,
\label{eq:gco}
\end{align}
where we define the unary term as:
\begin{align}
\Phi_i(L_i) \propto \varphi_i(z_i) 
\cdot \mu_i(x_i,z_i) \phi_i(x_i) 
\cdot \nu_i(y_i,z_i) \psi_i(y_i) 
\enspace,
\label{eq:gco_unary}
\end{align}
and the pairwise term as:
\begin{align}
&\Phi_{ij}(L_i,L_j) \propto \label{eq:gco_pairwise}\\
&\left \{\begin{array}{ll}
\phi_{ij}(x_i,x_j) \varphi_{ij}(z_i, z_j) 
& \mbox{if $L^o_i \neq L^o_j$ \& $L^a_i = L^a_j$}\\
\psi_{ij}(y_i,y_j) \varphi_{ij}(z_i, z_j) 
& \mbox{if $L^o_i = L^o_j$ \& $L^a_i \neq L^a_j$}\\
\phi_{ij}(x_i,x_j) \psi_{ij}(y_i,y_j) \varphi_{ij}(z_i, z_j) 
& \mbox{if $L^o_i \neq L^o_j$ \& $L^a_i \neq L^a_j$}\\
1 & \mbox{if $L^o_i = L^o_j$ \& $L^a_i = L^a_j$}
\enspace. \nonumber
\end{array} \right.
\end{align}
Figure \ref{fig:inference} shows a visualization example of the trilayer model after the rewritten. The edges linking $(x_i,z_i)$ and $(y_i,z_i)$ become a part of the unary potential in Eq.~\ref{eq:gco_unary}. The pairwise term as defined in Eq.~\ref{eq:gco_pairwise} satisfies the submodular property. Therefore, Eq.~\ref{eq:gco} can be efficiently optimized using $\alpha$-expansion or $\alpha \beta$-swap based multi-label graph cuts. 

% XXX Details of learning ... 

%--------------------------------------------------------------------
\section{Bridging the Gap Between Recognition and Segmentation}
\label{sec:holistic}

We have considered the video-level actor-action recognition in Sec.~\ref{sec:prob:single} and Sec.~\ref{sec:prob:multi}, and the pixel-level actor-action segmentation in Sec.~\ref{sec:prob:seg}. Apparently, the latter is a much harder task; this is also backed up in our experiments in Sec.~\ref{sec:exp:short}. Therefore, it makes sense to use the results obtained from the easier task, i.e., the video-level actor-action recognition, to guide the harder tasks, i.e., pixel-level actor-action segmentation. In this section, we describe a simple but effective approach, inspired by~\cite{DeOsIsIJCV2012} to make use of the recognition responses in optimizing segmentation. 

The idea is to model the responses from video-level recognition as label costs for pixel-level segmentation. Let us define $\Phi_{\mathcal{V}} (l)$ as the classification score, in the range of ${(0,1]}$, for a given actor-action label $l$ from video-level recognition, and $\Phi_{\mathcal{V}} (\cdot)$ can be any model considered at the video-level (see Sec.~\ref{sec:prob:single}). We extend the pixel-level trilayer segmentation formulation in Eq.~\ref{eq:gco}, such that: 
\begin{align}
&P(\mathbf{x},\mathbf{y},\mathbf{z}|\mathcal{V}) 
\doteq P(\mathbf{L}|\mathcal{V}) \label{eq:gco_labelcost} \\
&\quad = \prod_{i\in\mathcal{V}} P(L_i) 
\prod_{i\in\mathcal{V}} \prod_{j\in\mathcal{E}(i)} P(L_i,L_j)
\cdot P(\mathbf{L}) \nonumber \\
&\quad \propto  \prod_{i\in\mathcal{V}} \Phi_i(L_i)
\prod_{i\in\mathcal{V}} \prod_{j\in\mathcal{E}(i)} \Phi_{ij} (L_i, L_j)
\cdot \prod_{l\in\mathcal{L}} \delta_l (\mathbf{L}) \Phi_{\mathcal{V}} (l)
\enspace,
\nonumber
\end{align}
where $\delta_l (\mathbf{L})$ has the form:
\begin{align}
\delta_l (\mathbf{L}) = 
\left \{\begin{array}{ll}
1 & \mbox{if $\exists i \in \mathcal{V}: L_i = l$ }\\
1 / \Phi_{\mathcal{V}} (l) & \text{otherwise}
\enspace.
\end{array} \right.
\label{eq:gco_indicator}
\end{align}
In other words, $\delta_l (\mathbf{L}) \Phi_{\mathcal{V}} (l)$ represent a penalty term: when a certain actor-action label $l$ is presented in the labeling field, it has to pay a penalty that is proportional to its video-level classification score $\Phi_{\mathcal{V}} (l)$; when label $l$ is not presented, the penalty is canceled. 

The formulation in Eq.~\ref{eq:gco_labelcost} has two utilities. First, it enforces a compact video labeling that is parsimony to the number of different labels used to describe the video. In other words, we prefer the output video labeling to be as \textit{clean} as possible so that it can be unambiguously used in other applications, e.g., we do not want too many fragmented segment noises to confuse a real-time robotic system. Second, rather than uniformly penalizing over all labels, the amount of penalty is proportional to its video-level recognition response, i.e., less penalty if the actor-action label is supported at the video-level and more penalty if it is not. Therefore, the modeling achieves our goal of guiding harder task using relatively more reliable responses from easier task. 

Equation~\ref{eq:gco_labelcost} is described in the context of the trilayer segmentation model. It is trivial to extend to other segmentation models. For the na\"ive Bayes segmentation model, we use separate scores for actors and actions. For all other models in actor-action segmentation, we use the trilayer model from the video-level actor-action recognition for implementing $\Phi_{\mathcal{V}} (\cdot)$. Notice that Eq. \ref{eq:gco_labelcost} can also be efficiently solved by graph cuts inference following~\cite{DeOsIsIJCV2012}.

%--------------------------------------------------------------------
\section{Experiments on Short Videos}
\label{sec:exp:short}

We thoroughly study each of the instantiations of the actor-action understanding problem with the overarching goal of assessing if the joint modeling of actor and action improves performance over modeling each of them independently, despite the large space. We discuss the experimental results obtained on A2D short videos in this section and leave the discussion of experiments on long videos in Sec.~\ref{sec:exp:long}. We follow the training and testing splits discussed in Sec. \ref{sec:data}; for assigning a single-label to a short video to evaluate the single-label actor-action recognition, we choose the label associated with the query for which we searched and selected that video from YouTube. 

%--------------------------------------------------------------------
\subsection{Single-Label Actor-Action Recognition}
\label{sec:exp:single}

Following the typical action recognition setup, e.g.,~\cite{LaIJCV2005}, we use the dense trajectory features (trajectories, HoG, HoF, MBHx and MBHy)~\cite{WaKlScIJCV2013} and train a set of one-versus-all SVM models (with RBF-$\chi^2$ kernels from LIBSVM~\cite{ChLiTIST2011}) for the label sets of actors, actions and joint actor-action labels. Specifically, when training the \textit{eating} classifier, the other seven actions are negative examples; when we train the \textit{bird-eating} classifier, we use the 35 other actor-action labels as negative examples. Notice that for video-level actor-action recognition, we do not consider actors with \textit{none} action since they are not considered as a dominant action for videos. To evaluate the actor-action tuple for the na\"ive Bayes model, we first train and test actor and action classifiers independently, and then score them together (i.e., a video is correct if and only if both actor and action are correct). 

\begin{table}[t]
\centering
\caption{Actor-Action Recognition Evaluated in Three Settings}
\label{tab:exp_ar}
\resizebox{\linewidth}{!}{
\begin{tabular}{ r | c c c | c c c }
\hline
& \multicolumn{3}{ c |}{\textbf{Single-Label (Accuracy)}} & \multicolumn{3}{c}{\textbf{Multiple-Label (MAP)}} \\
\cline{2-7}			
\textbf{Model} & \textbf{Actor} & \textbf{Action} & \textbf{\textless A,A\textgreater} & \textbf{Actor} & \textbf{Action} & \textbf{\textless A,A\textgreater} \\
\hline
Na\"{i}ve Bayes & 70.51 & 74.40 & 56.17 & 76.85 & 78.29 & 60.13 \\
JointPS & 72.25 & 72.65 & 61.66 & 76.81 & 76.75 & 63.87  \\
Trilayer & \textbf{75.47} & \textbf{75.74} & \textbf{64.88} & \textbf{78.42} & \textbf{79.27} & \textbf{66.86} \\
\hline  
\end{tabular}}
\end{table}

Table~\ref{tab:exp_ar}-left shows the classification accuracy of the na\"ive Bayes, joint product space and trilayer models in terms of classifying actor labels alone, action labels alone and joint actor-action labels. We note that the scores are not directly comparable along the columns (e.g., the space of independent actors and actions is significantly smaller than that of actor-action tuples); the point of comparison is along the rows. We observe that the independent model outperforms the joint product space model in evaluating actions alone; for this, we suspect that the regularity across different actors for the same action is underexploited in the na\"ive Bayes model and exploited in the joint product space model, but that results in more inter-class overlap in the latter case. For example, a \textit{cat-running} and a \textit{dog-running} have both similar and different signatures in space-time: the na\"ive Bayes model does not need to distinguish between them; the joint product space model does, but its effort is not appreciated in evaluating actions alone ignoring the actors. Furthermore, we find that when we consider both actor and action in evaluation, it is clearly beneficial to jointly model them. This phenomenon occurs in all of our experiments. Finally, the trilayer model outperforms the other two models in terms of both individual actor or action tasks  as well as the joint actor-action task. The reason is that the trilayer model incorporates both types of relationships that are separately modeled in the na\"ive Bayes and joint product space models.

%--------------------------------------------------------------------
\subsection{Multiple-Label Actor-Action Recognition}
\label{sec:exp:multi}

For the multiple-label case, we use the same dense trajectory features as in Sec.~\ref{sec:exp:single}, and we train one-versus-all SVM models again for the label sets of actor, action and actor-action tuples, but with different training regimen to capture the multiple-label setting. For example, when training the \textit{adult} classifier, we use all videos containing actor \textit{adult} as positive samples no matter the other actors that coexist in the video, and we use the rest of videos as negative samples. For evaluation, we adapt the approach from HOHA2~\cite{MaLaScCVPR2009}. We treat multiple-label actor-action recognition as a retrieval problem and compute mean average precision (mAP) given the classifier scores.

Table~\ref{tab:exp_ar}-right shows the performance of the three methods on this task. Again, we observe that the joint product space model has higher mAP than the na\"ive Bayes model in the joint actor-action task, and lower mAP in individual tasks. We also observe that the trilayer model further improves the scores following the same trend as in the single-label recognition case. Furthermore, we observe large improvement in both individual tasks when comparing the trilayer model to the other two. This implies that the ``side'' information of the actor when doing action recognition (and vice versa) provides useful information to improve the inference task in a careful modeling, thereby answering the core question in the paper.  

%--------------------------------------------------------------------
\subsection{Actor-Action Segmentation}
\label{sec:exp:seg}

\begin{table*}[t]
\centering
\caption{Average Per-Class Accuracy for Actor-Action Segmentation Evaluated in Three Settings}
\label{tab:exp_seg_summary}
\resizebox{0.70\linewidth}{!}{
\begin{tabular}{ r | c c c | c c c | c c c }
\hline
& \multicolumn{3}{ c |}{\textbf{Unary}} & \multicolumn{3}{ c |}{\textbf{+ Pairwise}} & \multicolumn{3}{ c }{\textbf{+ Label Costs}}\\
\cline{2-10}
\textbf{Model} & \textbf{Actor} & \textbf{Action} & \textbf{\textless A,A\textgreater} & \textbf{Actor} & \textbf{Action} & \textbf{\textless A,A\textgreater} & \textbf{Actor} & \textbf{Action} & \textbf{\textless A,A\textgreater}\\
\hline
Na\"{i}ve Bayes & 43.02 & 40.08 & 16.35 & 44.78 & 42.59 & 19.30 & 44.37 & 46.14 & 22.00 \\
JointPS & 40.89 & 38.50 & 20.61 & 41.96 & 40.09 & 21.73 & 44.00 & 43.90 & 24.79 \\
Conditional & 43.02 & 41.19 & 22.55 & 44.78 & 41.88 & 24.19 & 44.37 & 41.85 & 24.12 \\
Bilayer & 43.02 & 40.08 & 16.35 & 44.46 & 43.62 & 23.43 & 45.98 & 48.31 & 27.89 \\
Trilayer & \textbf{43.08} & \textbf{41.61} & \textbf{22.59} & \textbf{45.70} & \textbf{46.96} & \textbf{26.46} & \textbf{47.39} & \textbf{49.39} & \textbf{30.53} \\
\hline  
\end{tabular}}
\end{table*}

\begin{table*}[t]
\centering
\caption{Segmentation Performance for Individual Actor-Action Tuples}
\label{tab:exp_seg_joint}
\resizebox{0.95\linewidth}{!}{
\begin{tabular}{ r | c c c c c c c c c c c c c c c }
\hline
& \multicolumn{1}{ c |}{} & \multicolumn{5}{ c |}{\textbf{baby}} & \multicolumn{4}{ c |}{\textbf{ball}} & \multicolumn{5}{ c }{\textbf{car}}\\
\cline{3-16}
\textbf{Model} & \multicolumn{1}{c|}{\textbf{BK}} & \textbf{climb} & \textbf{crawl} & \textbf{roll} & \textbf{walk} & \multicolumn{1}{c|}{\textbf{none}} & \textbf{fly} & \textbf{jump} & \textbf{roll} & \multicolumn{1}{c|}{\textbf{none}} & \textbf{fly} & \textbf{jump} & \textbf{roll} & \textbf{run} & \textbf{none} \\
\hline
Na\"{i}ve Bayes & \multicolumn{1}{c|}{82.75} & 19.93 & \textbf{25.36} & 23.78 & 12.81 & \multicolumn{1}{c|}{\textbf{14.59}} & 0.00 & 7.33 & 1.82 & \multicolumn{1}{c|}{0.00} & 16.80 & 74.19 & 27.98 & 11.87 & \textbf{16.32} \\
JointPS & \multicolumn{1}{c|}{83.26} & 2.53 & 17.89 & \textbf{52.56} & 14.98 & \multicolumn{1}{c|}{0.00} & 0.00 & 2.57 & 0.00 & \multicolumn{1}{c|}{0.00} & 13.10 & 83.75 & \textbf{51.73} & 26.01 & 7.38 \\
Conditional & \multicolumn{1}{c|}{82.75} & 14.86 & 17.63 & 36.77 & 12.00 & \multicolumn{1}{c|}{7.39} & 0.00 & \textbf{7.76} & 0.38 & \multicolumn{1}{c|}{0.00} & 24.96 & 79.22 & 41.58 & 19.51 & 0.00 \\
Bilayer & \multicolumn{1}{c|}{\textbf{85.73}} & \textbf{23.92} & 23.00 & 46.08 & \textbf{22.44} & \multicolumn{1}{c|}{0.00} & 0.00 & 6.03 & 0.12 & \multicolumn{1}{c|}{0.00} & 25.38 & 82.29 & 44.14 & 41.99 & 0.00 \\
Trilayer & \multicolumn{1}{c|}{83.44} & 14.90 & 24.01 & 46.70 & 15.81 & \multicolumn{1}{c|}{0.00} & 0.00 & 5.96 & \textbf{2.36} & \multicolumn{1}{c|}{0.00} & \textbf{37.81} & \textbf{83.77} & 50.90 & \textbf{51.33} & 0.00 \\
\hline \noalign{\medskip} \hline 
& \multicolumn{8}{ c |}{\textbf{adult}} & \multicolumn{7}{ c }{bird} \\
\cline{2-16}
\textbf{Model} & \textbf{climb} & \textbf{crawl} & \textbf{eat} & \textbf{jump} & \textbf{roll} & \textbf{run} & \textbf{walk} & \multicolumn{1}{c|}{\textbf{none}} & \textbf{climb} & \textbf{eat} & \textbf{fly} & \textbf{jump} & \textbf{roll} & \textbf{walk} & \textbf{none} \\
\hline
Na\"{i}ve Bayes & 33.26 & 49.87 & 38.76 & 26.48 & 16.73 & 15.90 & 17.99 & \multicolumn{1}{c|}{\textbf{19.99}} & \textbf{28.92} & 6.34 & 31.04 & 4.43 & 34.03 & 7.93 & \textbf{10.73} \\
JointPS & 24.75 & \textbf{66.30} & 62.37 & 33.18 & 18.03 & 38.13 & 22.94 & \multicolumn{1}{c|}{11.56} & 15.73 & \textbf{26.86} & 44.89 & 20.62 & 24.30 & 6.69 & 0.00 \\
Conditional & 21.54 & 52.78 & 49.05 & 27.91 & 30.76 & 29.04 & 21.16 & \multicolumn{1}{c|}{11.08} & 16.29 & 12.26 & 44.09 & 17.40 & 40.77 & 9.58 & 0.00 \\
Bilayer & 32.42 & 65.09 & \textbf{64.95} & 37.52 & \textbf{32.79} & 40.31 & 26.63 & \multicolumn{1}{c|}{10.46} & 18.15 & 15.39 & 49.09 & 18.91 & 29.22 & 15.30 & 0.00 \\
Trilayer & \textbf{41.29} & 65.25 & 58.99 & \textbf{42.33} & 28.36 & \textbf{51.15} & \textbf{35.61} & \multicolumn{1}{c|}{9.36} & 27.42 & 19.16 & \textbf{62.08} & \textbf{33.47} & \textbf{41.00} & \textbf{16.93} & 0.00 \\
\hline \noalign{\medskip} \hline 
& \multicolumn{7}{ c |}{\textbf{dog}} & \multicolumn{7}{ c |}{\textbf{cat}} & \\
\cline{2-15}
\textbf{Model} & \textbf{crawl} & \textbf{eat} & \textbf{jump} & \textbf{roll} & \textbf{run} & \textbf{walk} & \multicolumn{1}{c|}{\textbf{none}} & \textbf{climb} & \textbf{eat} & \textbf{jump} & \textbf{roll} & \textbf{run} & \textbf{walk} & \multicolumn{1}{c|}{\textbf{none}} & \textbf{Ave.} \\
\hline 
Na\"{i}ve Bayes & 11.05 & 21.61 & \textbf{8.91} & 26.81 & 23.98 & 36.95 & \multicolumn{1}{c|}{\textbf{1.85}} & 31.85 & 34.72 & 4.30 & 36.13 & 49.65 & 0.00 & \multicolumn{1}{c|}{\textbf{2.13}} & 22.00 \\
JointPS & 13.35 & 39.14 & 1.26 & 44.18 & 22.49 & 40.13 & \multicolumn{1}{c|}{0.00} & 17.01 & 33.91 & 12.50 & 40.24 & 35.17 & \textbf{19.15} & \multicolumn{1}{c|}{0.00} & 24.79 \\
Conditional & 6.40 & 22.71 & 6.69 & 25.95 & 17.65 & 52.16 & \multicolumn{1}{c|}{0.00} & 29.40 & 39.30 & 10.42 & 38.37 & \textbf{68.90} & 14.89 & \multicolumn{1}{c|}{0.00} & 24.12 \\
Bilayer & \textbf{22.16} & 25.36 & 0.05 & \textbf{49.23} & \textbf{26.88} & 44.21 & \multicolumn{1}{c|}{0.00} & 31.28 & \textbf{39.32} & \textbf{16.98} & 48.74 & 51.33 & 14.41 & \multicolumn{1}{c|}{0.00} & 27.89 \\
Trilayer & 20.37 & \textbf{41.80} & 2.09 & 39.21 & 24.81 & \textbf{52.25} & \multicolumn{1}{c|}{0.00} & \textbf{39.18} & 35.06 & 15.33 & \textbf{48.85} & 56.40 & 18.35 & \multicolumn{1}{c|}{0.00} & \textbf{30.53} \\
\hline
\end{tabular}}
\end{table*}

%--------------------------------------------------------------------
\subsubsection{Experiment Setup}

We use TSP~\cite{ChWeIICVPR2013} to obtain supervoxel segmentations due to its strong performance on the supervoxel benchmark~\cite{XuCoIJCV2016}. In our experiments, we set $k=400$ yielding about 400 supervoxels touching each frame. We compute histograms of textons and dense SIFT descriptors over each supervoxel volume, dilated by 10 pixels. We also compute color histograms in both RGB and HSV color spaces and dense optical flow histograms. We extract feature histograms from the entire supervoxel 3D volume, rather than a single representative superpixel~\cite{TiLaIJCV2012}. Furthermore, we inject the dense trajectory features~\cite{WaKlScIJCV2013} to supervoxels by assigning each trajectory to the supervoxels it intersects in the video.

Frames in A2D short videos are sparsely labeled; to obtain a supervoxel's ground-truth label, we look at all labeled frames in a video and take a majority vote over labeled pixels. We train one-versus-all SVM classifiers (linear kernels) for actors, actions, and actor-actions as well as conditional classifiers separately. The parameters of the graphical model are tuned by empirical search, and graph cut is used for inference as in Sec.~\ref{sec:prob:seg:infer}. The inference output is a dense labeling of video voxels in space-time, but, as our dataset is sparsely labeled in time, we compute the average per-class segmentation accuracy only against those frames for which we have ground-truth labels. We choose average per-class accuracy over global pixel accuracy because our goal is to compare the labelings of actors and actions other than background classes, which still dominates the majority of pixels in a video and all algorithms are performing quite well (all over 82\%, see the second column in Tab. \ref{tab:exp_seg_joint}).

%--------------------------------------------------------------------
\subsubsection{Benchmark State-of-The-Art CRFs}

Before we discuss the performance of segmentation models proposed in this paper, we benchmark two strong CRF models that are originally proposed for image segmentation. The first is the robust $P^N$ model~\cite{LaRuKoICCV2009} that defines a multi-label random field on a superpixel hierarchy and performs inference exhaustively from finer levels to coarser levels. We apply their supplied code off-the-shelf as a baseline. The average per-class accuracy is 13.7\% for the joint actor-action tuple, 47.2\% for actor alone and 34.49\% for action alone. We suspect that the modeling at pixel and superpixel levels can not well capture the motion changes of actions, which explains why the actor score is high but the other scores are comparatively lower. 

The second method is the fully-connected CRF~\cite{KrKoNIPS2011} that builds a pairwise dense random filed and imposes Gaussian mixture kernels to regularize the pairwise interactions. We extend their algorithm to use the same supervoxels and features as models considered in this paper and implement it based on the joint product space unaries. Its average per-class accuracy achieves 25.4\% for the joint actor-action tuple, 44.8\% for actor alone and 45.5\% for action alone. With the video-oriented features, the fully-connected CRF has surpassed many baseline segmentation models considered in this paper, e.g., the na\"{i}ve Bayes and joint product space models, but still has a clear gap to the trilayer model. This is expected as the dense edges have already modeled significant pairwise interactions but it  lacks the explicit consideration of various actor and action dependencies as in the trilayer model. 

%--------------------------------------------------------------------
\subsubsection{Results \& Discussion}

\begin{figure*}[t]
\centering
\includegraphics[width=\linewidth]{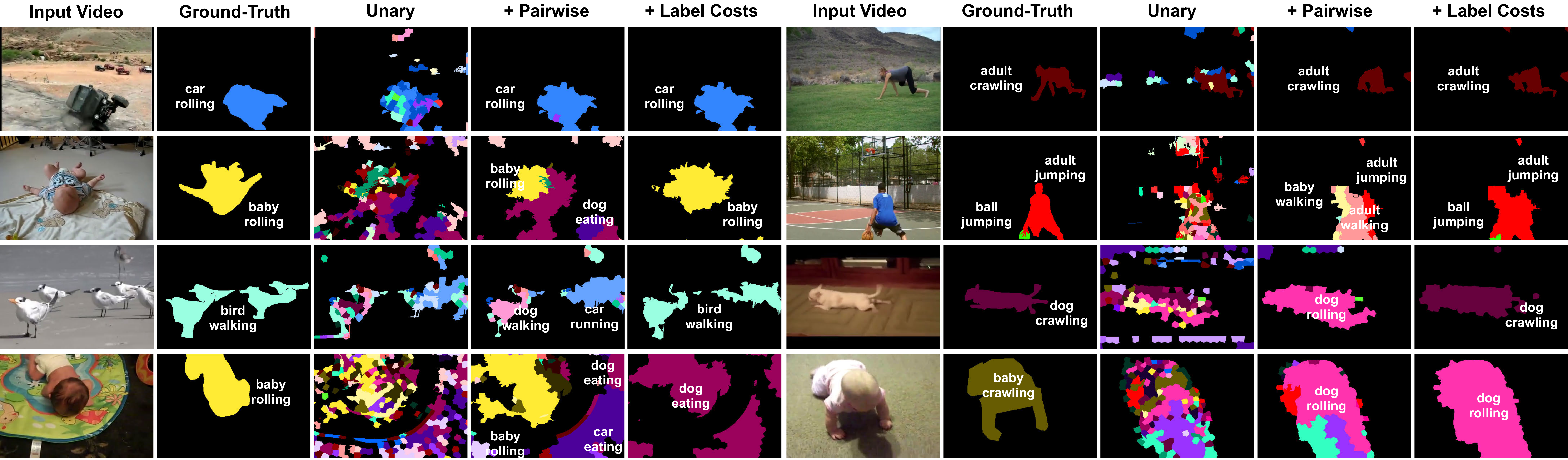}
\caption{Actor-action segmentation of the trilayer model in three settings: unary only, unary and pairwise interactions, and the full model with label costs. One frame per video is sampled. Best view zoomed and in color.}
\label{fig:visual_trilayer}
\end{figure*}

\noindent \textbf{Overall Performance.} \quad We first analyze the overall performance of various segmentation models in Table~\ref{tab:exp_seg_summary}. From left to right are results obtained with unary classifiers alone, unary and pairwise interactions, and full model with video-level recognition as label costs, respectively. We evaluate the actor-action tuples as well as individual actor and action tasks. Notice that the conditional model is a variation of bilayer model with different aggregation---we infer the actor label first then the action label conditioned on the actor. We also note that the bilayer model has the same unary scores as the na\"ive Bayes model (using actor $\phi_i$ and action $\psi_i$ outputs independently) and the actor unary of the conditional model is the same as that of the na\"ive Bayes model (followed by the conditional classifier for action). 

Although for the individual tasks, the na\"ive Bayes model is not the worst one, it performs worst when we consider the actor-action tuples, which is expected as it dose not encode any interactions between the two sets of labels. The joint product space model outperforms the na\"ive Bayes model for the actor-action tuples, but it has worse performance for the individual actor and action tasks, which is also observed in the actor-action recognition task. The conditional model has better actor-action scores, albeit it uses the inferred actors from the na\"ive Bayes model, which indicates that knowing actors can help with action inference. We also observe that the bilayer model has a poor unary performance of 16.35\% (for actor-action tuple) that is the same as the na\"ive Bayes model but after the edges between the actor and action nodes are added, it improves dramatically to 23.43\%, which suggests that the performance boost comes from the interactions of the actor and action nodes in the bilayer model. We also observe that the trilayer model has not only much better performance in the joint actor-action task, but also better scores for the individual actor and action tasks, as it is the only model that considers all three types of tasks together---the individual actor and action tasks, the joint space actor-action task and the conditional tasks. Furthermore, all models have received significant performance boost when the extra information from video-level recognition is used. 

\noindent \textbf{Individual Actor-Action Tuples.} \quad We compare the performance for individual actor-action tuples in Table~\ref{tab:exp_seg_joint}. The models considered here are full models with video-level recognition as label costs. We observe that the trilayer model has leading scores for more actor-action tuples than the other models, and the margin is significant for labels such as \textit{bird-flying} and \textit{adult-running}. We also observe a systematic increase in performance as more complex actor-action variable interactions are included. We note that the tuples with \textit{none} action are sampled with greater variation than the action classes (see Fig. \ref{fig:dat:color} for examples), which contributes to the poor performance of \textit{none} over all actors. Interestingly, the na\"ive Bayes model has relatively better performance on the \textit{none} action classes. We suspect that the label-variation for \textit{none} leads to high-entropy over its classifier density and hence when joint modeling, the actor inference pushes the action variable away from the \textit{none} action class.  

\begin{figure*}[t]
\centering
\includegraphics[width=\linewidth]{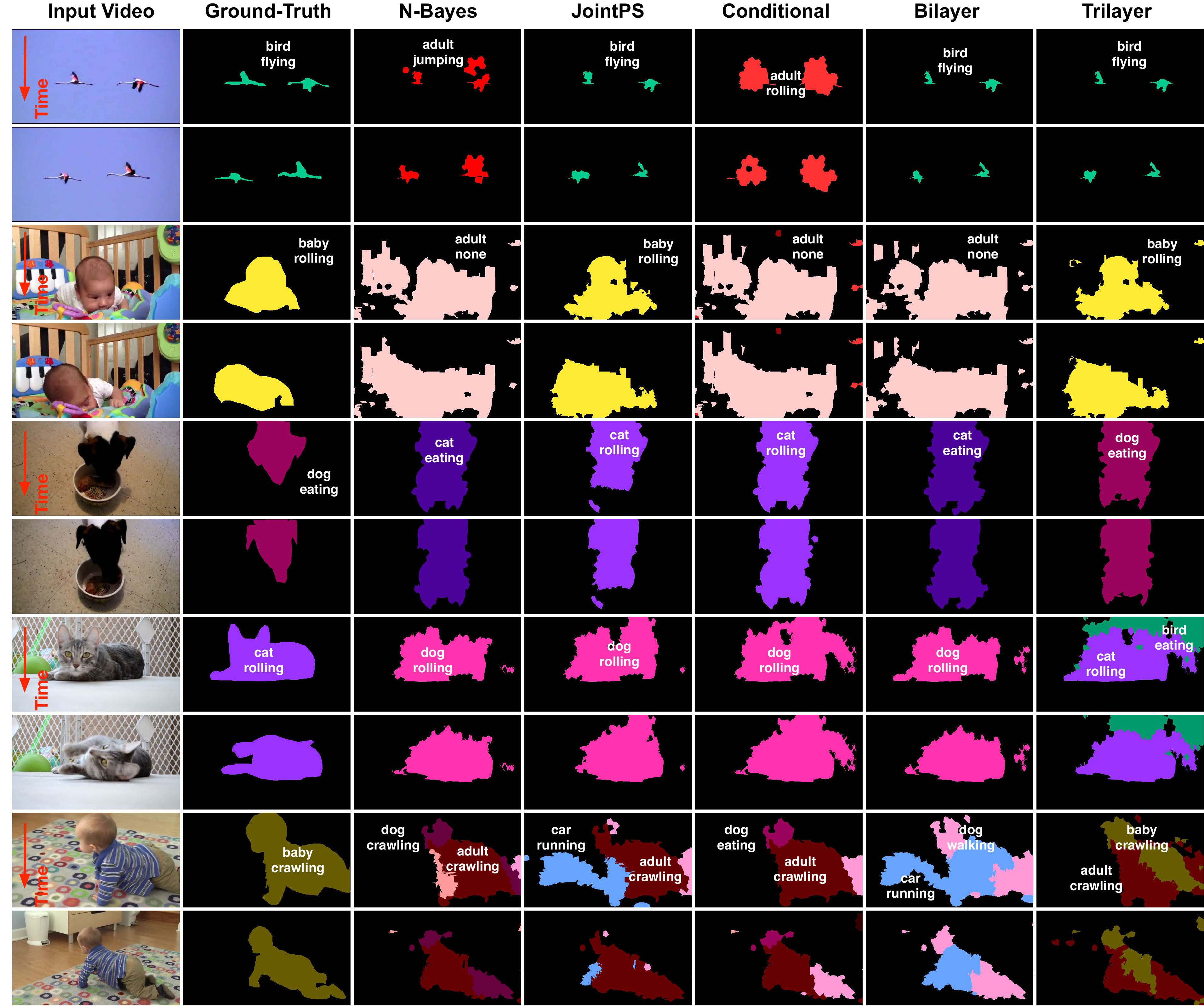}
\caption{Actor-action segmentation of all models when full model with label costs is considered. Best view zoomed and in color.}
\label{fig:visual_all_labelcost}
\end{figure*}

\noindent \textbf{Qualitative Results.} \quad We first show intermediate results leading to the full model of trilayer segmentation in Fig.~\ref{fig:visual_trilayer}, which include unary  responses alone, and unary responses and pairwise interactions. The full model contains unary responses, pairwise interactions and label costs from video-level recognition. The first row shows examples where both pairwise and full model are mostly correct. The next two rows contain examples where the video-level recognition via label costs help recover partially correct and even wrong segmentations produced by the pairwise model. Notice that our model differs from a uniform distribution of label costs, which can only enforce the compactness of segmentation and is hard to correct segmentations, where most of local predictions are wrong, e.g., the \textit{bird-walking} and \textit{dog-rolling} examples. One limitation of our model is that when video-level recognition generates poor predictions, it may also result in incorrect pixel-level segmentation, as shown in the last row of the figure. 

Finally, we show the comparison of all actor-action interaction models in Fig.~\ref{fig:visual_all_labelcost}, where they use the same video-level recognition via label costs. The point of comparison is at the various ways they model the interplay of actors and actions. Recall that the na\"ive Bayes model considers the actor and action tasks independent of each other; it gets many partially correct segmentations, e.g., the \textit{cat-eating} v.s. ground-truth \textit{dog-eating} and the \textit{dog-rolling} v.s. ground-truth \textit{cat-rolling}, but none of them agrees completely with ground-truth labels. The bilayer model adds a compatibility term between actors and actions, and correctly recovers the \textit{bird-flying} in the first video, while the conditional model fails since it infers action after inferring actor. The joint product space model correctly recovers the \textit{bird-flying} and the \textit{baby-rolling} in the first two videos. The trilayer model considers all situations in other models, and recovers the correct segmentations in most videos. The \textit{baby-crawling} example in the last video is particularly hard, where the trilayer model mislabels part of it as \textit{adult-crawling} suggesting the need of object-level awareness in the future work. 

%--------------------------------------------------------------------
\section{Experiments on Long Videos}
\label{sec:exp:long}

In this section, we evaluate our segmentation models on the 31 untrimmed long videos, which contain over 58K frames with dense annotations in time as introduced in Sec.~\ref{sec:dat:long}. We focus the evaluation on the temporal performance, e.g., the continuity of performing one action and the sensitivity of action changes, rather than the exact spatial location of segmentation as we evaluate on the short videos. This provides us a complementary viewpoint to assess our overarching goal, i.e., whether the joint modeling of actor and action improves performance over modeling each of them independently. 

%--------------------------------------------------------------------
\subsection{Evaluation Metric}

Despite the large actor-action label space, actors in A2D short videos only perform one type of actions throughout an entire video; this assumption certainly does not hold in reality: most of our queried YouTube videos contain actors performing a series of actions. Take the video shown in the top row of Fig.~\ref{fig:img:long} as an example: the blue-clothe baby performs \textit{walking}, \textit{rolling}, \textit{walking}, then \textit{none}; and the blue ball performs \textit{rolling}, \textit{jumping}, then \textit{none}. For videos like this, we have manually labeled bounding boxes on actors throughout the video and recorded the start and finish timestamps of actions we consider in the paper. We define an \textbf{actor-action track} as a tube composed of bounding boxes on actor that performs one type of actions for a continues period of time defined by the start and finish timestamps of the action. Therefore, in the previous example, the blue-clothe baby have four actor-action tracks and the blue ball has three actor-action tracks. There are a total of 727 actor-action tracks for A2D long videos (128 for validation and 599 for testing). Our evaluation metric is, indeed, designed to measure the recall of such actor-action tracks. 

Formally, let us define an actor-action track as $T = \{\mathbf{B}^{t_1}_l, \dots, \mathbf{B}^{t_m}_l\}$, where $\mathbf{B}^{t_i}_l$ denotes the bounding box on frame $t_i$ with actor-action label $l$, and $t_1$ and $t_m$ are the start and finish timestamps of the track. The actor-action video segmentation to be evaluated is denoted as a sequence of image segmentations $\mathbf{L} = \{\mathbf{L}^1, \dots, \mathbf{L}^n\}$, where $n$ is the total number of frames in video. The temporal recall of the actor-action track is defined as:
\begin{align}
R(\mathbf{L}, T) = \frac{1}{m} 
\sum_{i=1}^m \mathds{1}[ | \mathbf{L}^{t_i}(\mathbf{B}^{t_i}_l) | >0]
\enspace,
\label{eq:eval:long:r}
\end{align}
where $| \mathbf{L}^{t_i}(\mathbf{B}^{t_i}_l) |$ denotes the sum of pixels that match actor-action label $l$ within the segmentation region at $\mathbf{L}^{t_i}$ cropped by the bounding box $\mathbf{B}^{t_i}_l$. Intuitively, Eq.~\ref{eq:eval:long:r} measures how well in temporal domain the track is covered by segmentation, and it counts a match as long as the segmentation overlaps with bounding box region spatially in a frame.

We evaluate over all actor-action tracks $\{T_1, \dots, T_K\}$ in the testing set and we have:
\begin{align}
\text{Recall}(\sigma) = \frac{1}{K} 
\sum_{k=1}^K \mathds{1}[R(\mathbf{L}, T_k) \geq \sigma]
\enspace,
\end{align}
where $\sigma$ is a recall threshold meaning that we only count a positive recall if larger or equal to $\sigma$ of the track is covered by the correct segmentation. To generate a plot, we vary $\sigma$ from 0 to 1 by a step size of 0.1. Notice that we apply $\mathds{1}[R(\mathbf{L}, T_k) > \sigma]$ for $\sigma=0$. 

At the time of writing, there is not a well-established evaluation metric for measuring the quality of spatiotemporal segmentations in untrimmed long videos. The metric proposed in this section aims to measure the temporal performance of actor-action segmentation models. It is complementary to the spatial-oriented evaluation in Sec.~\ref{sec:exp:short} as the metric here does not consider how well the segmentation overlaps with ground-truth spatially. For a comprehensive understanding, both evaluations defined here and in Sec.~\ref{sec:exp:short} should be considered.

%--------------------------------------------------------------------
\subsection{Results \& Discussion}

\begin{figure}[t]
\centering
\includegraphics[width=0.80\linewidth]{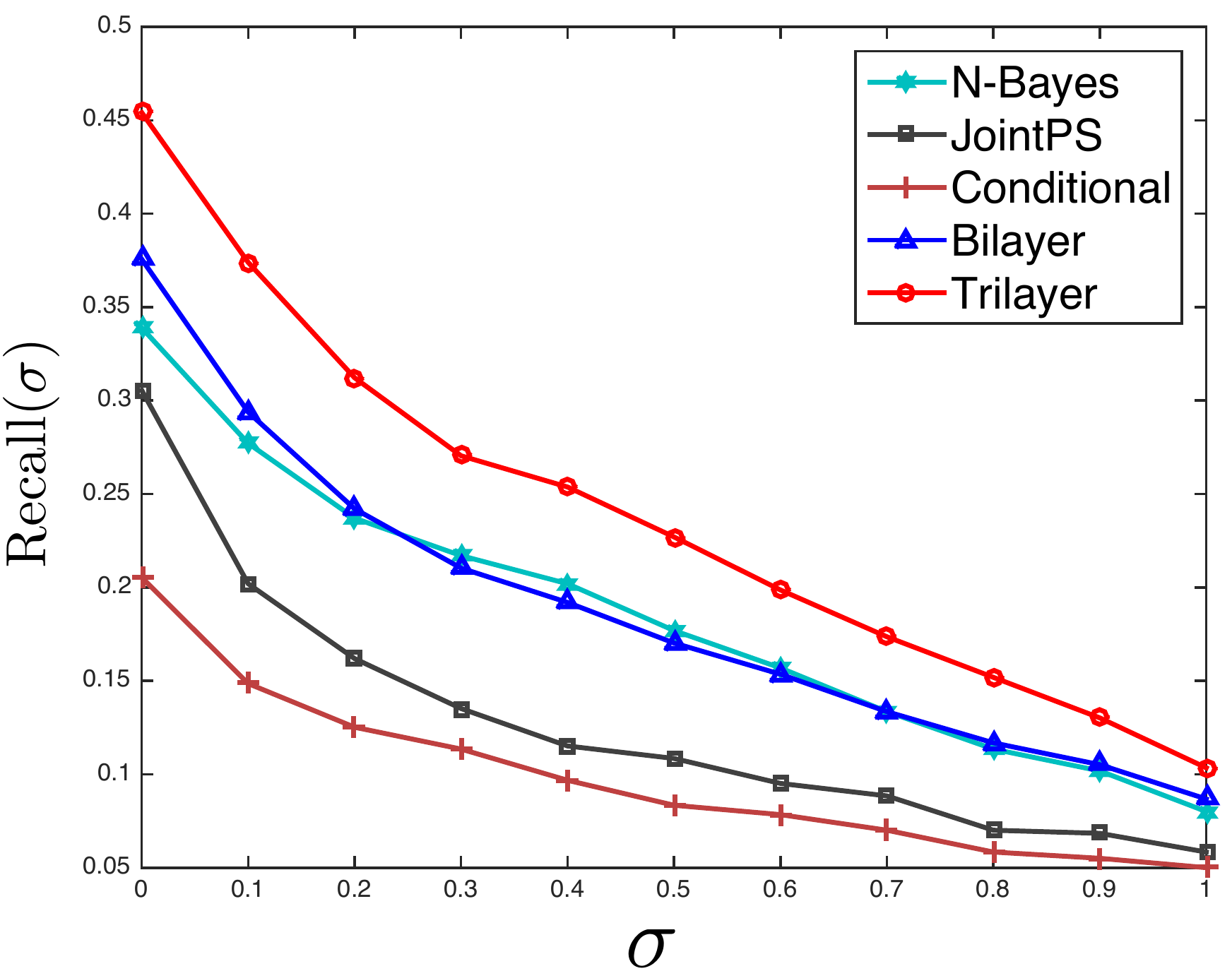}
\caption{The recall of actor-action tracks by varying $\sigma$.}
\label{fig:exp_long}
\end{figure}

We evaluate our actor-action segmentation models on long videos, where we only model unary and pairwise interactions as described in Sec.~\ref{sec:prob:seg}. We do not consider video-level actor-action recognition or its utility as label costs in segmentation, since the task is ill-posed in dealing with temporally untrimmed videos where a single actor may perform multiple different actions. Again, we use TSP~\cite{ChWeIICVPR2013} to obtain supervoxel segmentations and compute the same set of features as we do for short videos. Since we have limited number of long videos, the unary classifiers are trained on the training set of short videos where ground-truth segmentations are available; the parameters are empirically tuned using validation set of long videos.  

The recall plot of the actor-action tracks for all segmentation models is shown in Fig.~\ref{fig:exp_long}. In general, the recall decreases when $\sigma$ increases; this is expected as the measure of positive recall is getting more strict as $\sigma$ increases. The trilayer segmentation model outperforms other models by a significant margin. It achieves a 45\% recall of all tracks when $\sigma=0$. Even in the extreme case, when $\sigma=1$ meaning that the track region at every frame is covered by at least one correct segment, the trilayer model still get roughly 10\% of all actor-action tracks correct. 

For other models, the bilayer model outperforms the joint product space and conditional models. Interestingly, the na\"{i}ve Bayes model performs as good as the bilayer model for $\sigma>0.2$. We suspect that the noisy segmentations produced by the independent modeling of actors and actions have tricked our evaluation metric when the spatial overlapping of segments is not strictly examined. Furthermore, when combined with the spatial-oriented evaluation in Table~\ref{tab:exp_seg_summary}, it is clear that the bilayer model outperforms the na\"{i}ve Bayes model. 

In summary, the results we obtained on long videos have confirmed our findings from short videos that the explicitly joint modeling of actors and actions outperforms modeling them independently. We also note that the current methods have difficulties in dealing with untrimmed long videos as seen from relatively poor performance in general. We have released the newly collected long videos along with their annotations and evaluation metric in A2D v2 and look forward to seeing future methods toward solving the challenges of segmenting untrimmed videos.

%--------------------------------------------------------------------
\section{Conclusion}
\label{sec:con}

Our thorough assessment of all instantiations of the actor-action understanding problem on short videos and actor-action segmentation on long videos provides strong evidence that the joint modeling of actor and action improves performance over modeling each of them independently. We find that for both individual actor and action understanding and joint actor-action understanding, it is beneficial to jointly consider actor and action. A proper modeling of the interactions between actor and action results in dramatic improvement over the baseline models of the na\"ive Bayes and joint product space models, as we observe from the bilayer and trilayer models. 

Our paper set out with three goals: first, we sought to motivate and develop a new, more challenging, and more relevant actor-action understanding problem; second, we sought to assess whether joint modeling of actors and actions improves performance for this new problem using both short and long videos; third, we sought to explore a multi-scale modeling to bridge the gap between recognition and segmentation tasks. 
We achieved these goals through the following four contributions:
\begin{enumerate}[labelsep=5pt, labelwidth=0pt,leftmargin=12pt,itemsep=0ex, 
parsep=0pt, topsep=0pt, partopsep=0pt]
\item New actor-action understanding problem and fully labeled dataset of 3782 short videos and 31 long videos.  
\item Thorough evaluation of actor-action recognition and segmentation problems using state-of-the-art features and models. The experiments unilaterally demonstrate a benefit for jointly modeling actors and actions.
\item A new trilayer model that combines independent actor and action variations and product-space interactions.
\item Improved the performance of pixel-level segmentation via label costs from video-level recognition responses. 
\end{enumerate}
Our full dataset, computed features, codebase, and evaluation regimen are released at \url{http://www.cs.rochester.edu/~cxu22/a2d/} to support further inquiries into this new and important problem in video understanding.

\noindent \textbf{Limitations.} \quad There are two directions for our future work. First, our models lack the concept of modeling object as a whole entity; this results in fragmented segments in our actor-action segmentation. It would be beneficial to incorporate such concept in modeling actors and actions to achieve a holistic video understanding. Second, deep learning and deep representations are emerging topics in video understanding. It is our future work to explore the actor-action problem in these contexts.

\ifCLASSOPTIONcompsoc
  \section*{Acknowledgments}
\else
  \section*{Acknowledgment}
\fi

This work has been supported in part by Google, Samsung, DARPA W32P4Q-15-C-0070 and ARO W911NF-15-1-0354.

\ifCLASSOPTIONcaptionsoff
  \newpage
\fi



\bibliographystyle{IEEEtran}
\bibliography{IEEEabrv,A2D}
% \end{thebibliography}

% \begin{IEEEbiography}[{\includegraphics[width=1in,height=1.25in,clip,keepaspectratio]{mshell}}]{Michael Shell}

\begin{IEEEbiography}[{\includegraphics[width=1in,height=1.25in,clip,keepaspectratio]{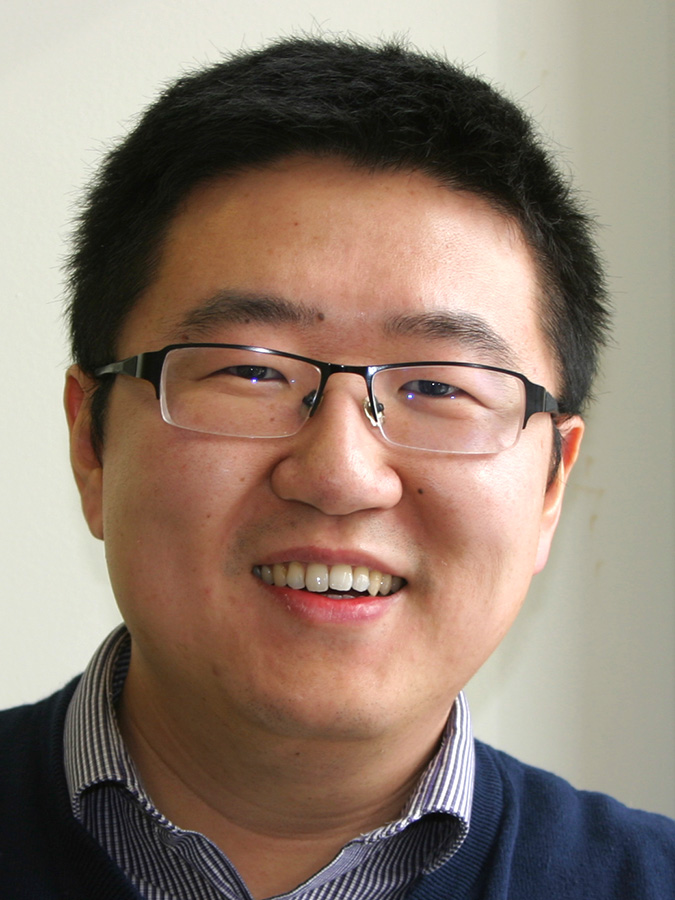}}]{Chenliang Xu} 
is an assistant professor of Computer Science at the University of Rochester. He received his Ph.D. degree at the University of Michigan, Ann Arbor in 2016, and M.S. degree from SUNY Buffalo in 2012, both in Computer Science. He received his B.S. degree in Information and Computing Science from Nanjing University of Aeronautics and Astronautics in 2010. His research interests include computer vision, robot perception, and machine learning.
\end{IEEEbiography}

\begin{IEEEbiography}[{\includegraphics[width=1in,height=1.25in,clip,keepaspectratio]{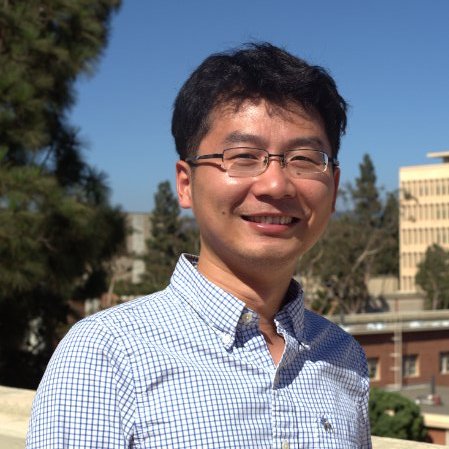}}]{Caiming Xiong}
is a senior researcher at Salesforce MetaMind. Before that, he was Postdoctoral Researcher in the Department of Statistics, University of California, Los Angeles. He received a Ph.D. in the Department of Computer Science and Engineering, SUNY Buffalo in 2014, and got the B.S. and M.S. of Computer Science degree from Huazhong University of Science and Technology in the year 2005 and 2007 in China. His research interests include video understanding, action recognition, metric learning and active clustering, and human-robot interaction.
\end{IEEEbiography}

\begin{IEEEbiography}[{\includegraphics[width=1in,height=1.25in,clip,keepaspectratio]{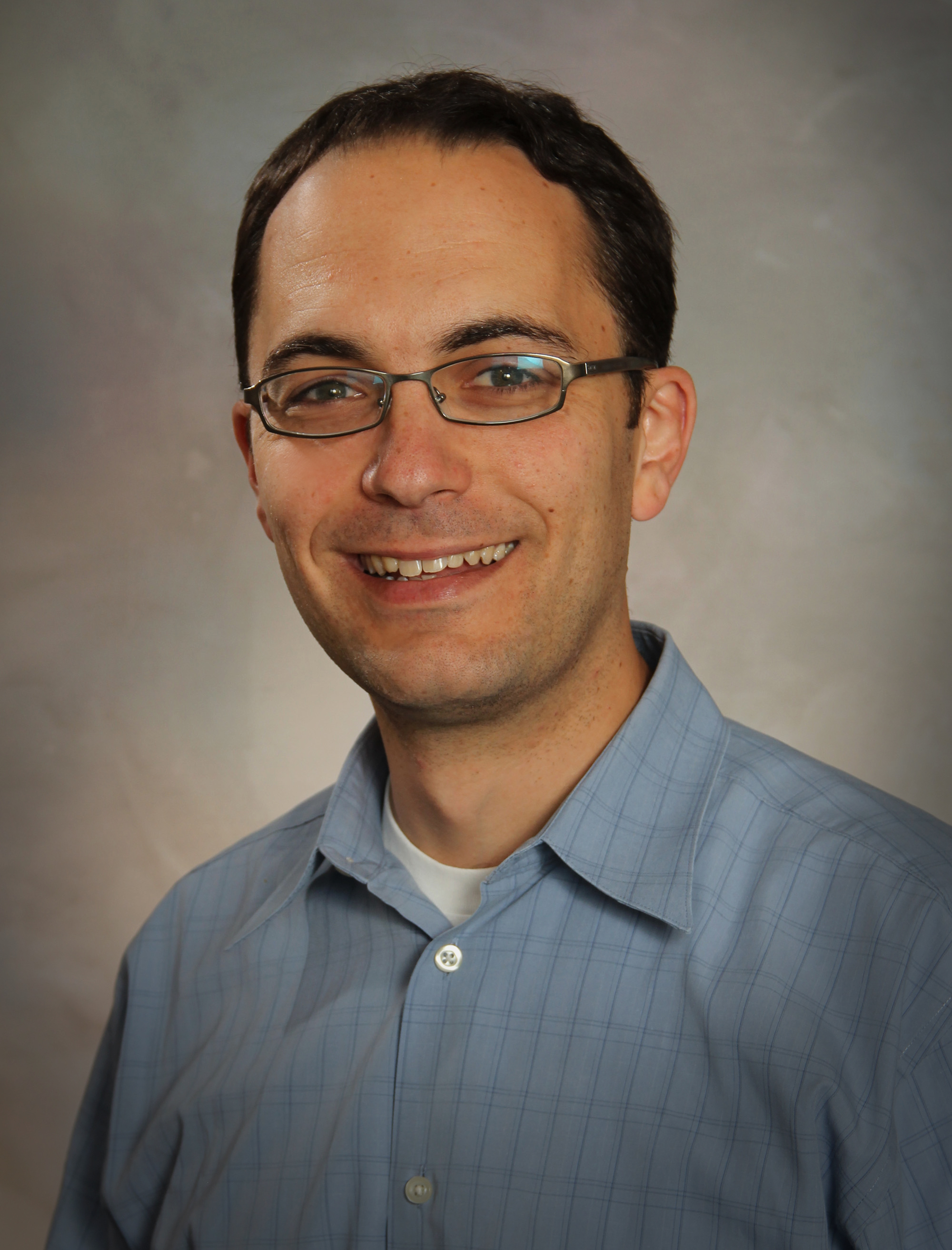}}]{Jason J. Corso} is an associate professor of Electrical Engineering and Computer Science at the University of Michigan.  He received his PhD and MSE degrees at The Johns Hopkins University in 2005 and 2002, respectively, and the BS Degree with honors from Loyola College In Maryland in 2000, all in Computer Science.  He spent two years as a post-doctoral fellow at the University of California, Los Angeles. From 2007-14 he was a member of the Computer Science and Engineering faculty at SUNY Buffalo.  He is the recipient of a Google Faculty Research Award 2015, the Army Research Office Young Investigator Award 2010, NSF CAREER award 2009, SUNY Buffalo Young Investigator Award 2011, a member of the 2009 DARPA Computer Science Study Group, and a recipient of the Link Foundation Fellowship in Advanced Simulation and Training 2003.  Corso has authored more than ninety peer-reviewed papers on topics of his research interest including computer vision, robot perception, data science, and medical imaging.  He is a member of the AAAI, IEEE and the ACM.
\end{IEEEbiography}


%\newpage



%\vfill


\end{document}